\pdfoutput=1
\documentclass[11pt]{article}

\usepackage[]{EMNLP2023}

\usepackage{times}
\usepackage{latexsym}

\usepackage[T1]{fontenc}
\usepackage[utf8]{inputenc}

\usepackage{microtype}

\usepackage{inconsolata}

\usepackage{graphicx}
\usepackage{subfigure}

\usepackage{amsmath}
\usepackage{amstext}
\usepackage{amsfonts}
\usepackage{bm}

\usepackage{bbm}
\usepackage{multirow}
\usepackage{booktabs}
\usepackage{array}
\usepackage{colortbl}
\usepackage{color}
\definecolor{mygray}{gray}{.9}
\definecolor{c0}{cmyk}{1,0.3968,0,0.2588} 

\usepackage{soul}

\usepackage{algorithm}
\usepackage{algorithmic}
\usepackage{todonotes}

\usepackage{amssymb}  
\usepackage{pifont}

\newcommand{\greenyes}{\textcolor{green}{\ding{51}}}
\newcommand{\redno}{\textcolor{red}{\ding{55}}}

\usepackage{enumitem}
\setenumerate[1]{itemsep=0pt,partopsep=0pt,parsep=\parskip,topsep=5pt}
\setitemize[1]{itemsep=0pt,partopsep=0pt,parsep=\parskip,topsep=5pt}
\setdescription{itemsep=0pt,partopsep=0pt,parsep=\parskip,topsep=5pt}

\usepackage{tcolorbox}
\newtcbox{\hlprimarytab}{on line, box align=base, colback=c0!10,colframe=white,size=fbox,arc=3pt, before upper=\strut, top=-2pt, bottom=-4pt, left=-2pt, right=-2pt, boxrule=0pt}

\newtcbox{\hlsecondarytab}{on line, box align=base, colback=c1!10,colframe=white,size=fbox,arc=3pt, before upper=\strut, top=-2pt, bottom=-4pt, left=-2pt, right=-2pt, boxrule=0pt}

\definecolor{c0}{cmyk}{1,0.3968,0,0.2588} 
\definecolor{c1}{cmyk}{0,0.6175,0.8848,0.1490} 
\newcommand{\dashifted}{{$\downarrow$}} % \tiny
\newcommand{\da}[1]{{\hlprimarytab{\dashifted{#1}}}} % \scriptsize
\newcommand{\uashifted}{{$\uparrow$}}
\newcommand{\ua}[1]{{\hlsecondarytab{\uashifted{#1}}}}

\usepackage{xspace}
\newcommand{\method}{\textsc{ImageNetVC}\xspace}

\title{\method: Zero- and Few-Shot Visual Commonsense Evaluation \\ on 1000 ImageNet Categories}

\author{{Heming Xia}\textsuperscript{\rm 1,\rm 2}\thanks{\ \ Co-first authors with equal contributions}, {Qingxiu Dong}\textsuperscript{\rm 1,\rm 3}\footnotemark[1], {Lei Li}\textsuperscript{\rm 1,\rm 3}, {Jingjing Xu}\textsuperscript{\rm 4}, \\
{\textbf{Tianyu Liu}}\textsuperscript{\rm 5}, {\textbf{Ziwei Qin}}\textsuperscript{\rm 1,\rm 3}, {\textbf{Zhifang Sui}}\textsuperscript{\rm 1,\rm 3}\\
  \textsuperscript{\rm 1}National Key Laboratory for Multimedia Information Processing, Peking University\\
  \textsuperscript{\rm 2}School of Software \& Microelectronics, Peking University \\
  \textsuperscript{\rm 3}School of Computer Science, Peking University \quad \textsuperscript{\rm 4}Shanghai AI Lab\quad \textsuperscript{\rm 5}Alibaba\\ 
  {\tt \{xiaheming,szf\}@pku.edu.cn; \{dqx,qinziwei\}@stu.pku.edu.cn}
}

\begin{document}
\maketitle
\begin{abstract}
Recently, Large Language Models (LLMs) have been serving as general-purpose interfaces, posing a significant demand for comprehensive visual knowledge.
However, it remains unclear how well current LLMs and their visually augmented counterparts (VaLMs) can master visual commonsense knowledge. 
To investigate this, we propose \method, a human-annotated dataset specifically designed for zero- and few-shot visual commonsense evaluation across 1,000 ImageNet categories. 
Utilizing \method, we benchmark the fundamental visual commonsense knowledge of both unimodal LLMs and VaLMs.
Furthermore, we analyze the factors affecting the visual commonsense knowledge of large-scale models, providing insights into the development of language models enriched with visual commonsense knowledge. Our code and dataset are available at \url{https://github.com/hemingkx/ImageNetVC}.
\end{abstract}

\section{Introduction}
With the breakthrough progress of Large Language Models (LLMs) in recent years~\cite{Brown:2020gpt3,opt}, LLMs are gradually adopted as general-purpose API interfaces (e.g., ChatGPT\footnote{\url{https://chat.openai.com}}). 
In addition to language, these intelligent agents, are further required to understand vision knowledge~\cite{hao2022language}, especially the visual perception, which is crucial for real-world interactions such as commonsense reasoning \cite{Talmor:2019commonsenseQA}, recipe generation \cite{Agarwal:2020recipe}, and robotic navigation \cite{Shah:2022navigation}.

\begin{figure}[t]
\centering
\includegraphics[width=0.9\columnwidth]{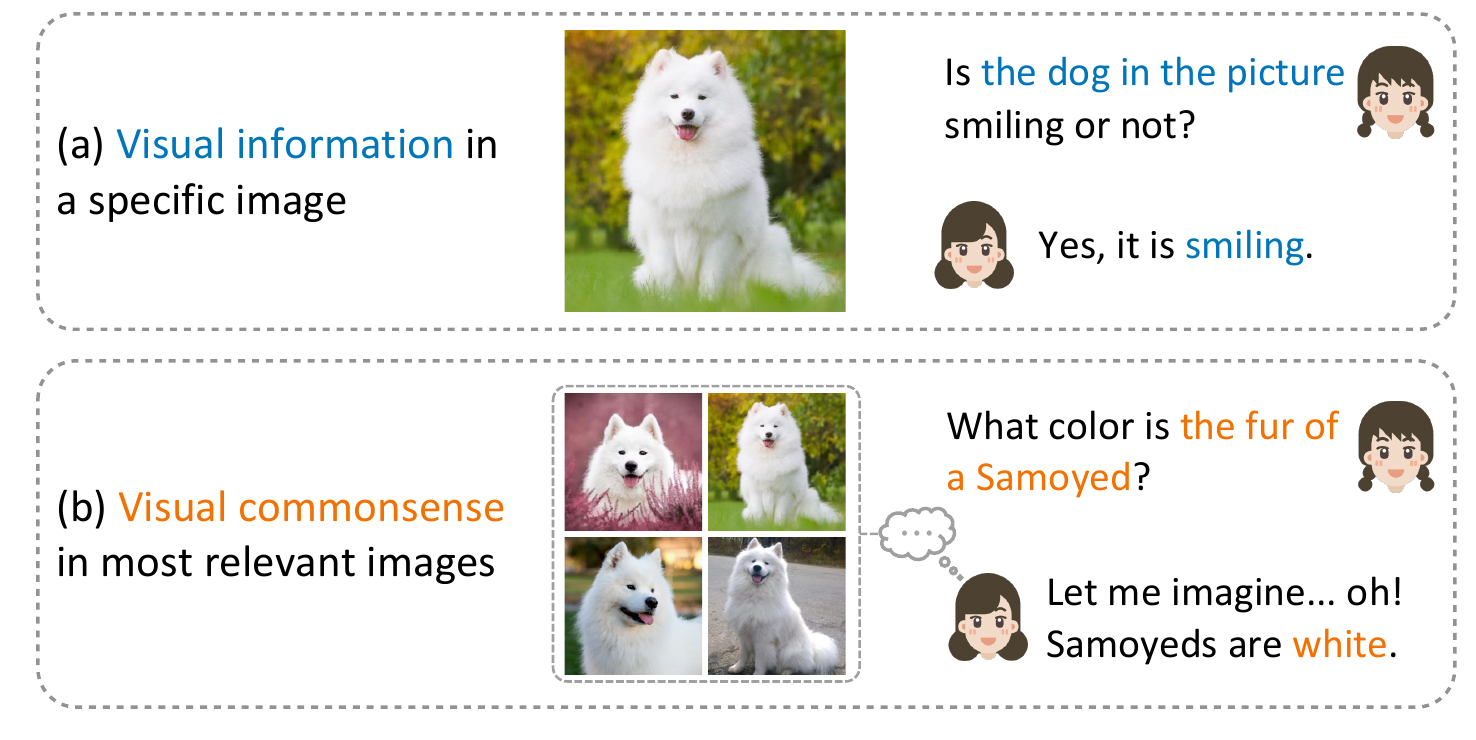}
\caption{Illustration of Visual Commonsense. Visual commonsense refers to the general visual knowledge that is commonly shared across the world, as opposed to the visual information that is specific to a single image. Visual Commonsense can be captured through a series of related images.}
\label{fig:VisualCommonsense}
\end{figure}

However, current studies lack a systematic evaluation on how well these widely-used LLMs and their variants are capable of visual understanding. Recent research proposes to evaluate the visual capability of models through visual commonsense evaluation~\cite{relativesize,memorycolor}. As shown in Figure~\ref{fig:VisualCommonsense}, visual commonsense evaluation aims to evaluate the model's understanding of commonly shared human knowledge about generic visual concepts, including color~\cite{Bruni:2012colorterm, memorycolor, Zhang:2022vicomte}, spatial relations~\cite{Liu:2022spatialcs}, relative sizes~\cite{relativesize}, etc. Despite their insightful investigations, these studies still have the following limitations from two sides: 1) \textbf{data side}: some research mines visual commonsense attributes based on frequency distributions in plain text corpora, which diverges from human visual perception and exhibits additional textual bias~\cite{Zhang:2022vicomte}; 2) \textbf{model side}: most existing evaluations only focus on a specific model group, lacking a comprehensive exploration of various model families~\cite{relativesize, memorycolor, Liu:2022spatialcs}.

In this work, we propose that similar to human beings, models can also answer intricate visual commonsense questions with related images (illustrated in Figure~\ref{fig:VisualCommonsense}). 
To this end, we introduce \method, a unified zero- and few-shot visual commonsense benchmark incorporating multiple sources of images (e.g., ImageNet~\cite{imagenet}, search images, and synthetic images). 
From the data side, \method comprises 4,076 high-quality QA pairs, encompassing 1,000 ImageNet categories across diverse domains such as color, shape, material, component, etc. 
Moreover, as a human-annotated dataset, \method utilizes human visual perception to identify shared attributes across relevant images, avoiding textual bias and providing data that is more closely aligned with human knowledge. From the model side, besides unimodal LLMs, \method also enables the evaluation of various Visually-augmented Language Models (VaLMs) to investigate the effect of visual grounding, which compensates for the lack of images in previous benchmarks.

With \method, we conduct extensive evaluations on the most widely-used LLMs and VaLMs. 
We benchmark the visual commonsense capabilities of various LLMs such as OPT, LLaMA, and Falcon and assess the effect of visual grounding in VaLMs with multiple sources of relevant images. We further analyze the co-founding factors that may affect the visual commonsense capability of models, such as model scale, in-context learning, and image sources. 
We highlight several experimental findings. These findings support the high value of our benchmark in assessing visual commonsense capabilities.

\begin{itemize}
    \item Template-based datasets yield artificially inflated and unstable visual commonsense evaluation, while our manually constructed \method provides evidence that visual commonsense remains challenging for LLMs.
    \item We discover that the acquisition of visual commonsense is an emergent ability for LLMs. For instance, 1.3B could be a potential threshold for unimodal LLMs to emergent with visual commonsense on the component.
    \item In-context learning enhances the understanding of visual commonsense tasks for both LLMs and VaLMs, not only reducing their variance across prompts but also calibrating the model confidence on visual commonsense.
\end{itemize}

\begin{figure*}[t]
\centering
\includegraphics[width=0.95\textwidth]{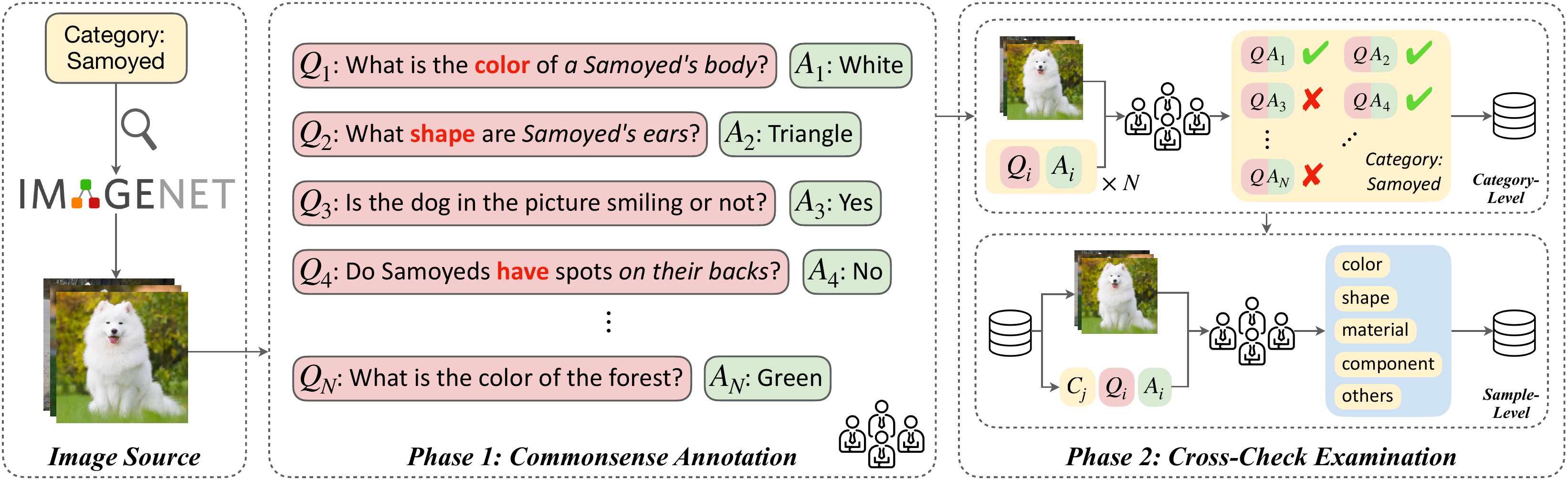}
\caption{An overall demonstration of the construction procedures of \method.}
\label{fig:ImageNetVC}
\end{figure*}
\begin{table*}[t]
    \centering
    \small
    \setlength{\tabcolsep}{3pt}
    \begin{tabular}{@{}lcccccrr@{}}
    \toprule
      \bf Dataset & \begin{tabular}[c]{@{}c@{}}Human \\Annotation\end{tabular} & \begin{tabular}[c]{@{}c@{}}Multi-attribute \\Evaluation\end{tabular} & \begin{tabular}[c]{@{}c@{}}Support \\VaLM\end{tabular} & \begin{tabular}[c]{@{}c@{}}Region-based \\Question\end{tabular} & \begin{tabular}[c]{@{}c@{}}Natural \\Language\end{tabular} & \#Category &\#Test \\
      \midrule
      MemoryColors~\cite{memorycolor}  &\greenyes  & \redno & \greenyes & \redno & \redno & 109 &109\\
      ColorTerms~\cite{Bruni:2012colorterm} & \greenyes &\redno & \redno & \redno & \redno & 52 & 52\\
      RelativeSize~\cite{relativesize} & \greenyes & \redno & \redno & \redno & \redno & 41 & 486\\
      SpatialCS~\cite{Liu:2022spatialcs} & \greenyes & \redno & \redno & \redno & \redno & 59 & 1224\\
      ViComTe~\cite{Zhang:2022vicomte}  & \redno & \greenyes & \redno & \redno & \redno & 3957 & 2223  \\
      \rowcolor{c0!5} \method~(Ours) &  \greenyes & \greenyes & \greenyes & \greenyes & \greenyes & 1000 & 4076\\
    \bottomrule
    \end{tabular}
    \caption{Features and statistical information of ImageNetVC and prior related datasets. The `\# Category' column indicates the number of object categories included, and`\# Test' means the number of test samples in the dataset.}
    \label{tab:dataset-comparison}
\end{table*}

\section{Related Work}

\paragraph{Large Language Models}
Text-only Large language models (LLMs) have exhibited outstanding performance across various textual commonsense tasks, benefiting from their training on extensive textual data~\cite{Radford:2019gpt2,Raffel:2020t5,Brown:2020gpt3}. However, the lack of visual data (e.g., images) during pretraining restricts their visual commonsense capabilities~\cite{lilei2023}. On the other hand, Visually-augmented Language Models (VaLMs) have gained popularity by integrating visual information into LLMs~\cite{Tsimpoukelli:2021frozen, alayrac2022flamingo}, which enhance the visual understanding capabilities of language models~\cite{Yang:2022zlavi,wang2022visually}.

\paragraph{Visual Commonsense Evaluation}
Visual commonsense knowledge of visual concepts is a fundamental and critical aspect of AI systems seeking to comprehend and reason about the world~\cite{yao2022visually, pmr}. 
Previously, several datasets have been proposed to address specific attributes of visual commonsense, including MemoryColors~\cite{memorycolor}, ColorTerms~\cite{Bruni:2012colorterm}, RelativeSize~\cite{relativesize}, and Spatial Commonsense (SpatialCS)~\cite{Liu:2022spatialcs}. To evaluate general visual commonsense, \citet{Zhang:2022vicomte} introduced ViComTe, a template-based dataset consisting of various (subject, object) pairs (such as \emph{(sky, blue)}).
However, its reliance on pure textual input underestimates the visual capabilities of VaLMs. Furthermore, its utilization of template-based formats and automatic extraction techniques leads to substandard data quality and inherent textual biases.

In this work, we introduce \method, a human-annotated visual commonsense evaluation dataset that consists of 4K natural language QA pairs across various visual attributes, which supports both LLM and VaLM evaluation with multiple sources of images. We present detailed comparisons of \method with prior work in Table~\ref{tab:dataset-comparison}.

\section{\method}
Starting from ImageNet, we construct our \method dataset in a multi-step crowd-sourcing pipeline, including 1) annotator training, 2) commonsense annotation, and 3) cross-check examination. An overall demonstration of our annotation process is illustrated in Figure~\ref{fig:ImageNetVC}. 

\subsection{Image Source}
We selected ImageNet~\cite{imagenet} as our image source because it covers a large number of commonly used objects in real-life situations, providing a diverse and representative image source. Additionally, the unified image format in ImageNet with dimensions of 256$\times$256 pixels facilitates annotators' understanding of images and reduces feature engineering. Specifically, we used the widely-used ImageNet (ILSVRC) 2012 subset,\footnote{\url{image-net.org/challenges/LSVRC/2012/}} consisting of 1.4 million images from 1,000 object categories.

\subsection{Prerequisite: Annotator Training}
We posted online job listings on Amazon Mechanical Turk\footnote{\url{https://www.mturk.com/}} and received over 500 applications from candidates with Bachelor's degrees or higher. To ensure dataset quality, we provided training with instructions and guidelines and a quick quiz to assess candidate understanding. Only candidates with scores larger than 95\% are hired.

\subsection{Phase 1: Commonsense Annotation}
Figure~\ref{fig:ImageNetVC} shows the commonsense annotation phase, where annotators are provided with category names and 50 randomly sampled images per category. They are instructed to form a question and answer considering shared visual features of the images and their own commonsense knowledge. Visual features may be object-based, such as the color of a entire object, or region-based, such as the color of a specific object part. Annotators first identify a common visual feature of the category, such as \emph{The color of a Samoyed's body is white.} They then create a QA pair based on this feature if it aligns with their commonsense understanding of life, such as \emph{What is the color of a Samoyed's body? White.}

To ensure that the QA pairs reflect visual commonsense rather than visual information tailored to specific images, annotators are instructed to focus on the visual features of each category rather than individual images. They are also provided with annotation examples and guidelines for rejection. The annotation UI and specifications for annotation can be found in Appendix~\ref{appendix:annotation}.

\subsection{Phase 2: Cross-Check Examination}
The primary objective of the cross-check examination phase is to conduct a rigorous screening and categorization of high-quality QA pairs that meet our requirements. This phase comprises two stages. 
In Stage 1, a category-level examination is performed, where three examiners are assigned to all annotated QA pairs in the same category. They are required to check all the pairs in the category based on the annotation instructions, rectify any grammatical errors, and eliminate low-quality or non-compliant pairs. Only the QA pairs that all three examiners approve are deemed acceptable.
Stage 2 involves a sample-level examination. Although the examination in Stage 1 is efficient, examining all QAs in one category simultaneously creates a misalignment with the final testing method (one-by-one QA) and introduces a distribution bias in the examination process. Therefore, in Stage 2, a more thorough sample-level examination is carried out. Three examiners are randomly assigned a QA pair with the corresponding category name from the entire dataset. They vote on whether to accept the QA pair and classify it into the following five subsets: color, shape, material, component, and others. Only the sample that receives the majority vote is approved for acceptance.

\begin{table}[t]
    \centering
    \small
    \setlength{\tabcolsep}{2pt}
    \begin{tabular}{lrrrrrr}
\toprule
                 & Color&Shape&Mater.&Compo.&Others&Total \\
\midrule
  \bf \# Labels & 11  & 12  &   16   &    2    &  55  & 91   \\
 \bf \# Categories & 439 & 367 &  405   &   560   & 768  &1000  \\
\bf \# Samples     & 557 & 424 &  430   &  1114   & 1551 &4076  \\
\bottomrule
    \end{tabular}
    \caption{Statistical information of \method. \textit{Mater.} and \textit{Compo.} are the abbreviations of Material and Component, respectively.}
    \label{tab:dataset_info}
\end{table}

\begin{figure}[t]
\centering
\includegraphics[width=0.45\textwidth]{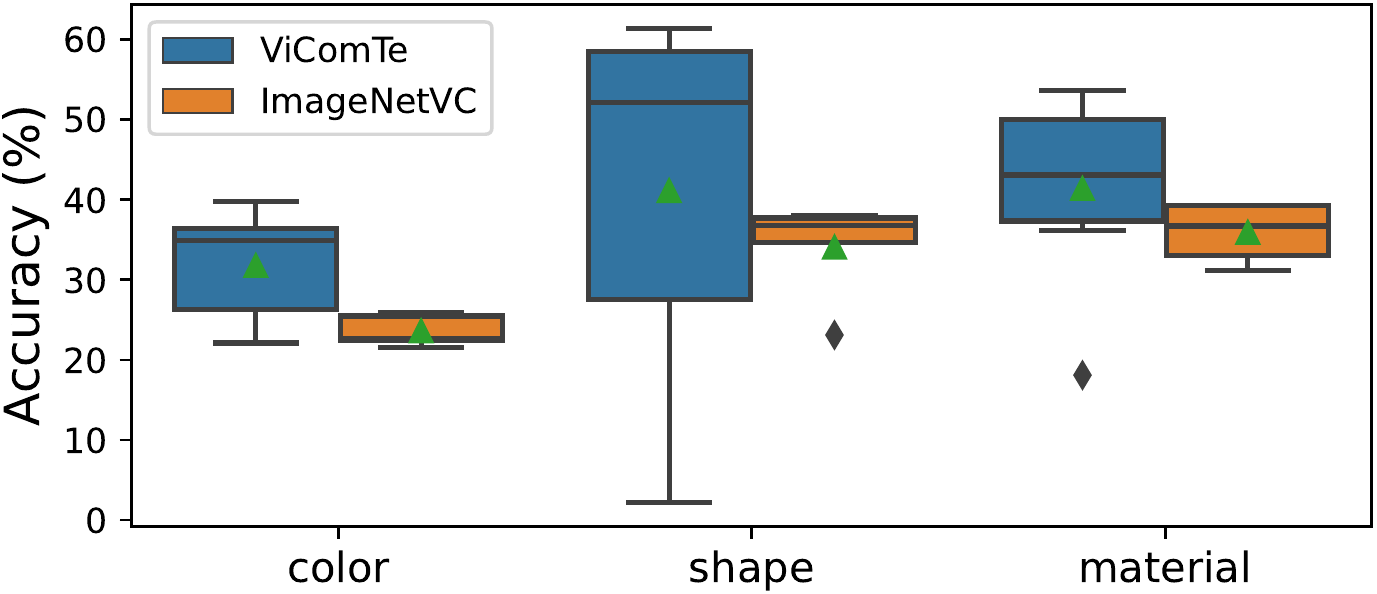}
\caption{Zero-shot performance distribution of GPT-Neo-1.3B across different prompts. Green triangles represent mean results, black diamonds represent outliers. Compared to the template-based dataset, ViComTe, \method demonstrates notably reduced evaluation variance across various prompts.}
\label{fig:box_vicomte}
\end{figure}

Our 60-day annotated dataset comprises 4,076 items (refer to Table~\ref{tab:dataset_info}) from 1000 ImageNet categories. It consists of 5 individual sub-tasks: color, shape, material, component, and others. More information and examples of \method can be found in Appendix~\ref{sec:details-of-imagenetvc}. All pricing strategy details and a hierarchical supervision process employed are elaborated in Appendix~\ref{sec:annotation-supervision} and \ref{pricing-strategy}.

\begin{figure}[t]
\centering
\includegraphics[width=0.45\textwidth]{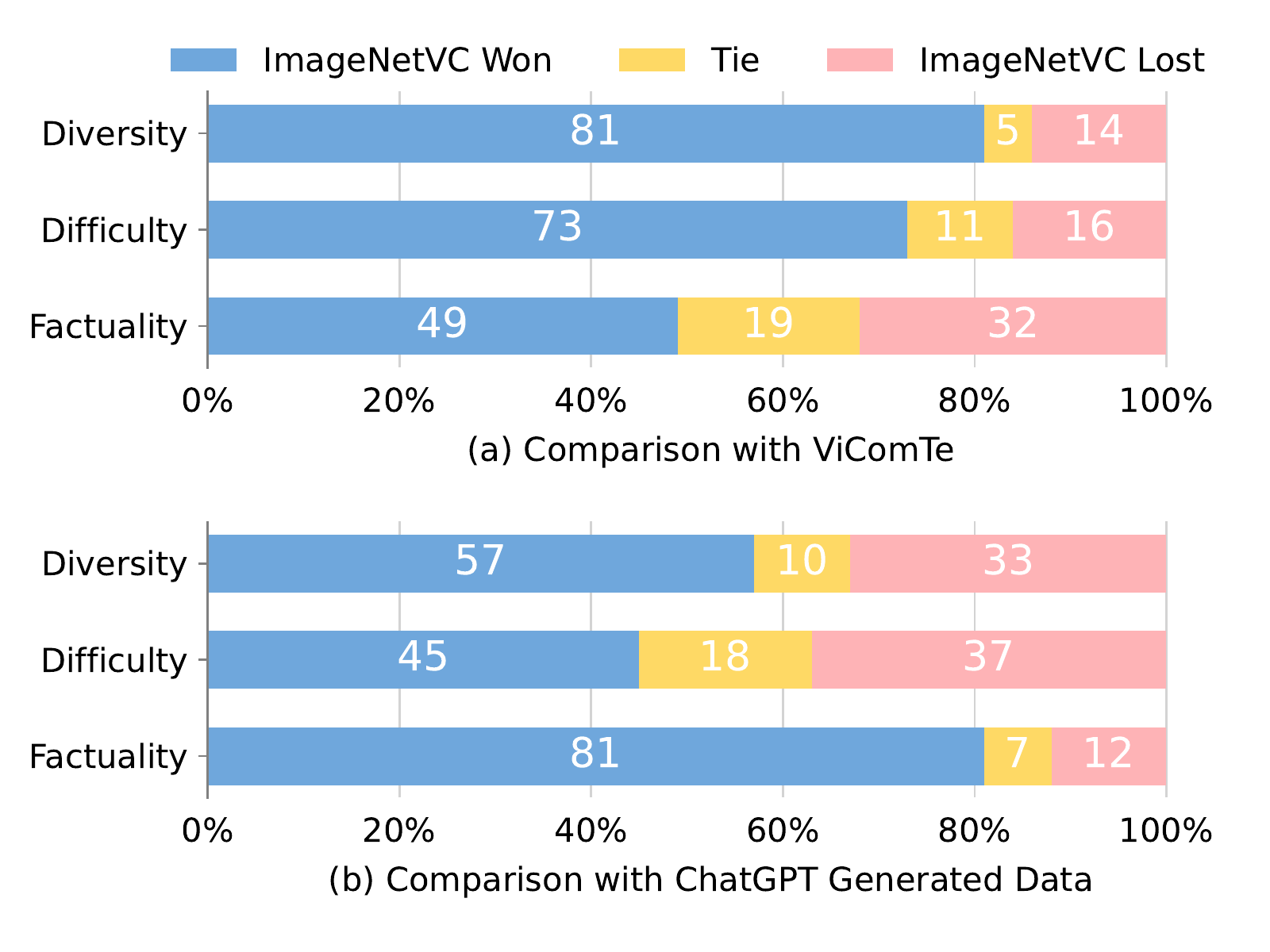}
\caption{Human assessment of visual commonsense dataset from three aspects: diversity, difficulty, and factuality. \method outperforms ViComTe in terms of diversity and difficulty, while also demonstrating superior factuality compared to ChatGPT generated data.}
\label{fig:quality_eval}
\end{figure}

\subsection{Dataset Evaluation}
Unlike previous datasets which are template-based, \method comes from diverse real images associated with human-annotated descriptions, which can better represent real-world settings. To assess the strengths of our dataset, we conduct automatic evaluation and human evaluation in this section.

First, we implement GPT-Neo-1.3B~\cite{gpt-neo} with respective subsets of \method and ViComTe, a widely-used dataset, across different prompts.\footnote{Prompt details are provided in Appendix~\ref{sec:eva-prompts}.} 
Results in Figure~\ref{fig:box_vicomte} indicate that, as a template-based dataset, ViComTe exhibits severe prompt bias, with substantial evaluation variance across different prompts. E.g., the model achieves only 2\% accuracy with the prompt ``X is of shape Y'' but achieves 61\% score with ``The shape of the X is Y''. Besides, compared with ViComTe, \method containing region-based questions is more challenging to models. For example, with the suitably selected prompt on the color subset, the model achieves 40\% accuracy on ViComTe but only 28\% accuracy on \method.

We further conducted a human assessment between ViComTe, \method, and QA pairs automatically generated from ChatGPT.\footnote{Please refer to Appendix~\ref{app:human-assess} for the detailed process of our designed human assessment.}  Specifically, we provided human annotators with sampled data from the two comparison datasets and asked them to vote for the better one considering diversity, difficulty, and factuality. As depicted in Figure~\ref{fig:quality_eval}, in more than 84\% of cases, \method outperforms or matches the template-based ViComTe in terms of diversity and difficulty, which is consistent with the results in Figure~\ref{fig:box_vicomte}. Moreover, our dataset demonstrates notably higher factual correctness compared to the data automatically generated by ChatGPT in more than 81\% of cases.

To sum up, our data collection process ensured high-quality annotations with minimal bias and increased diversity, difficulty, and factuality, providing a challenging dataset for advancing research in visual commonsense understanding.

\section{Experiments}
% \subsection{Models}
Our experiments primarily focus on two types of language models: LLMs and VaLMs. Both models have demonstrated promising capabilities in understanding visual information~\cite{lilei2023}.

\paragraph{Large Language Models} We begin with the text-only setting, to explore the visual commonsense capability of LLMs learned from the textual corpus. 
We focus on the dominant auto-regressive LLM family and benchmark a lot of model variants, including the GPT~\cite{gpt-neo, gpt-j, gpt-neox-20b}, OPT~\cite{opt}, LLAMA~\cite{llama}, Falcon~\cite{falcon}, and Pythia~\cite{pythia}.

\paragraph{Visually-augmented Language Models}
In our experiments, we mainly evaluate three widely-used open-source VaLMs: Z-LaVI~\cite{Yang:2022zlavi}, BLIP-2~\cite{Li:2023blip2}, and MAGMA~\cite{Eichenberg:2022magma}. These VaLMs are mainly built on top of \textit{frozen} LLMs and incorporate diverse mechanisms to integrate visual information. Further model details are provided in Appendix~\ref{sec:valm-intro}.

\subsection{Evaluation Methods}
In this work, we focus on evaluating the zero- and few-shot visual commonsense of LLMs and VaLMs on \method. Following \citet{Schick:2021pet} and \citet{Yang:2022zlavi}, we treat the zero-shot evaluation as a cloze test, transforming the QA pairs in \method into prompts like ``[Question] The answer is [Answer].''\footnote{All the prompts utilized for the evaluation of LLMs and VaLMs are shown in Appendix~\ref{sec:eva-prompts}.}. Formally, each QA pair is converted into a sequence of tokens $\bm{x} = \{x_0, ..., x_i, ..., x_n\}$, in which $x_i$ is the answer. In the few-shot setting, examples with the same prompt are concatenated before each QA pair.

\subsubsection{LLM Evaluation} 
Given an LLM $\mathcal{M}$, the sequence of input tokens $\bm{x} = \{x_0, ..., x_n\}$ will first be mapped to 
text embeddings $\bm{e}_{t}=\{\bm{e}_{t}(x_0), ..., \bm{e}_{t}(x_i), ..., \bm{e}_{t}(x_n)\}$ by the embedding layer $\bm{e}_{t} \in \mathcal{M}$. Then we utilize the model to calculate the score for the answer $y \in \mathcal{Y}$:
\begin{equation}
\small
\nonumber
\begin{aligned}
s_{t}(y \mid \bm{x})&=\frac{1}{-\log P_{\mathcal{M}}\left(x_{i} \mid \bm{x}_{<i}\right)} \\
&=\frac{1}{-\log P_{\mathcal{M}^{\prime}}\left(\bm{e}_{t}(x_{i}) \mid \bm{e}_{t}(\bm{x}_{<i})\right)}
\end{aligned}
\end{equation}
where $\mathcal{M}^{\prime}$ denotes the transformer neural network in $\mathcal{M}$, $P(\cdot)$ is the output probability given by the model. Then we obtain a probability distribution over all answer candidates using softmax:
\begin{equation}
\small
\label{eq:score-over-candidates}
q_{t}(y \mid \bm{x})=\frac{e^{s(y \mid \bm{x})}}{\sum_{y^{\prime} \in \mathcal{Y}} e^{s\left(y^{\prime} \mid \bm{x^{\prime}}\right)}}
\end{equation}
We calibrate the prediction by normalizing the probability distribution following~\citet{Zhao:2021calibratebeforeuse}, to mitigate the bias introduced by prompt formats as well as few-shot examples.
\begin{figure*}[t]
\centering
\includegraphics[width=0.95\textwidth]{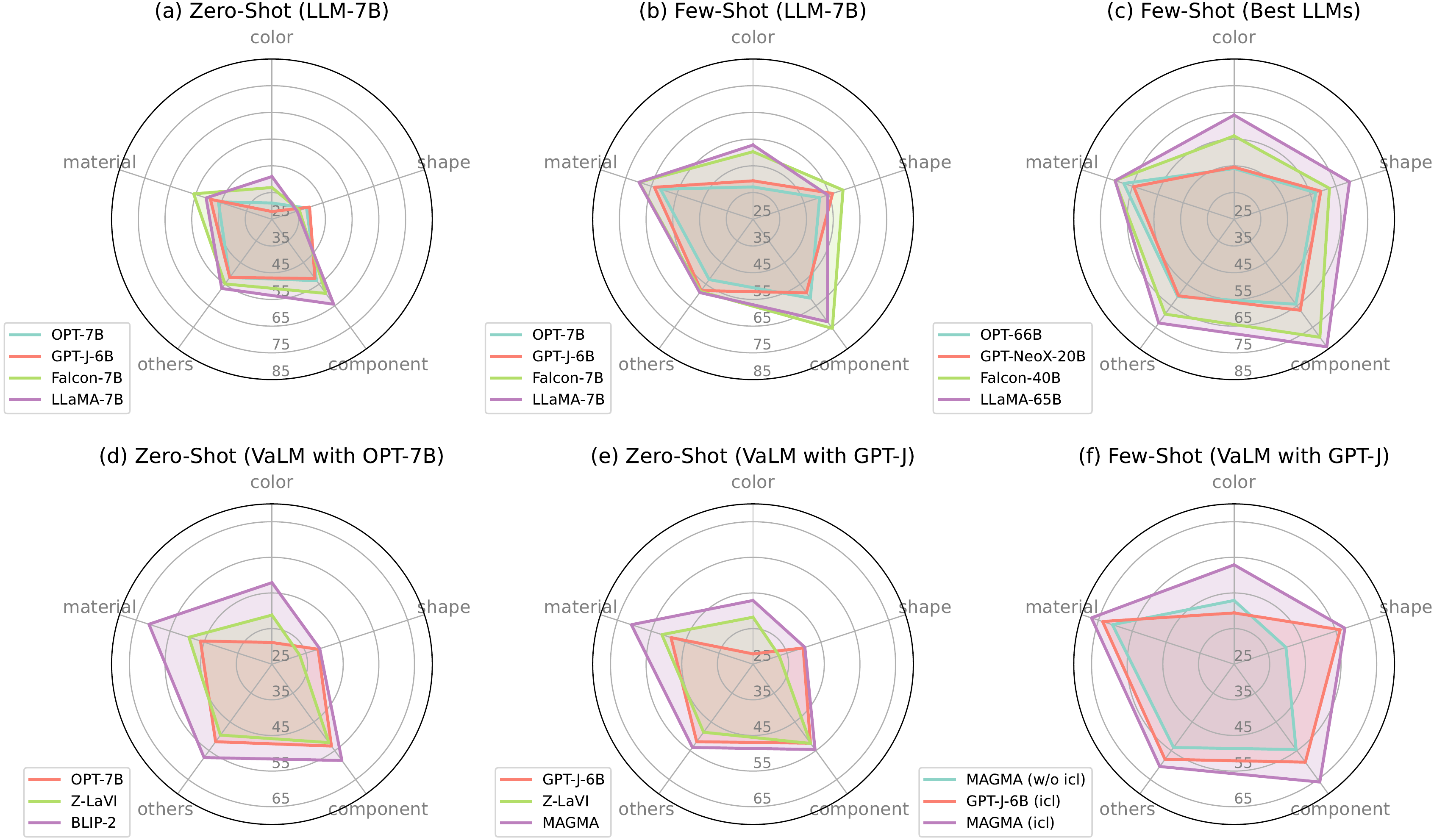}
\caption{Radar plots for five individual sub-tasks in \method. We show evaluation results with four experimental settings: \textbf{(a, b)} Zero- and few-shot evaluation with LLMs-7B; \textbf{(c)} Few-shot evaluation with the best LLMs in their own model family; \textbf{(d, e)} Zero-shot evaluation with VaLMs and their \textit{frozen} LLM backbones; \textbf{(f)} Few-shot evaluation with VaLMs. The numbers along the radio axis denote the mean Top-1 accuracy (\%) of models over 5 different prompts. The detailed results for drawing these plots are shown in Appendix~\ref{sec:main-results-details}.}
\label{fig:radar}
\end{figure*}
\subsubsection{VaLM Evaluation} 
\label{sec:valm-eval}
We incorporate two types of image sources as additional visual inputs for evaluating VaLMs: images retrieved from the web and synthetic images. We adopt Google Image Search to retrieve relevant images and Stable Diffusion~\cite{Rombach:2O22stablediffusion} for image synthesis. Following \citet{Yang:2022zlavi}, for each QA pair, we utilize CLIP~\cite{Radford:2021clip} to sort images from these two sources based on their similarity with the question and then preserve top-$K$ images as the final image sources. We mainly evaluate two types of VaLMs: \textit{prefix-based VaLMs} and \textit{ensemble-based VaLMs}. 

\paragraph{Prefix-based VaLMs}
Given a QA pair with an image $\bm{v}$, \textit{prefix-based VaLMs} (e.g., BLIP-2 and MAGMA) first utilize a visual encoder to transform the image into a sequence of visual embeddings $\bm{e}_{v} = \{\bm{e}_{v}^{1}, ..., \bm{e}_{v}^{m}\}$. Then, these embeddings are prefixed into the text embeddings of $\bm{x}$ and put into the \textit{frozen} LLM backbone to calculate the score:
\begin{equation}
\nonumber
\small
\begin{aligned}
s(y \mid \bm{v}, \bm{x})&=\frac{1}{-\log P_{\mathcal{M}}\left(x_{i} \mid \bm{v}, \bm{x}_{<i}\right)} \\
&=\frac{1}{-\log P_{\mathcal{M}^{\prime}}\left(\bm{e}_{t}(x_{i}) \mid \bm{e}_{v}, \bm{e}_{t}(\bm{x}_{<i})\right)}
\end{aligned}
\end{equation}
The probability distribution with the image $\bm{v}$ is calculated over all answer candidates, which is the same as Eq (\ref{eq:score-over-candidates}). If $K$ images are provided, the final distribution will be averaged over all images:
\begin{equation}
\small
\label{eq:prob-over-images}
q(y \mid \bm{v}, \bm{x})=\frac{1}{K} \sum_{i=1}^K q(y \mid \bm{v}^{(i)}, \bm{x})
\end{equation}
%following a widely used in-context learning (ICL) setting~\cite{icl}.

Following \citet{icl}, for \textit{prefix-based VaLMs} supporting few-shot evaluations, examples $\{\bm{v}^{(j)}, \bm{x}^{(j)}\}_{j=1}^{L}$ with the same processing will be concatenated in front of each QA pair.

Since \textit{prefix-based VaLMs} utilize \textit{frozen} LLM backbones, they can be regarded as a conditional extension of text-only LLMs. Evaluations between these two model types facilitate a thorough assessment of the effect of visual grounding on visual commonsense ability.

\paragraph{Ensemble-based VaLMs}
Given the input tokens $\bm{x}$ and multiple images $\bm{v}=\{\bm{v}^{(i)}\}_{i=1}^{K}$, \textit{ensemble-based VaLMs} (e.g., Z-LaVI) utilize a frozen CLIP model, which contains a text encoder $f_{t}$ and a visual encoder $f_{v}$, to project the tokens $\bm{x}$ and the image $\bm{v}^{(i)}$ into a shared representation space and compute the relevance score between them:

\begin{small}
\begin{equation}
\nonumber
s_{v}(y \mid \bm{v}^{(i)}, \bm{x})= \cos \left(f_{t}(\bm{x}), f_{v}(\bm{v}^{(i)})\right)
\end{equation}
\end{small}
Then, same as Eq (\ref{eq:score-over-candidates}) and Eq (\ref{eq:prob-over-images}), the probability distribution over all answer candidates and across $K$ images is obtained:

\begin{small}
\begin{equation}
\nonumber
q_{v}(y \mid \bm{v}, \bm{x})=\frac{1}{K} \sum_{i=1}^K\operatorname{softmax}\left(s_{v}(y \mid \bm{v}^{(i)}, \bm{x})\right)
\end{equation}
\end{small}
where $\operatorname{softmax}(\cdot)$ is a simplified denotation of Eq (\ref{eq:score-over-candidates}). The final ensemble score is calculated as a weighted sum over the output distributions of LLMs and CLIP:

\begin{small}
\begin{equation}
\nonumber
q(y \mid \bm{v}, \bm{x})=(1-w)\cdot q_{t}(y \mid \bm{x})+w\cdot q_{v}(y \mid \bm{v}, \bm{x})
\end{equation}
\end{small}
where $w$ denotes the weight hyperparameter.
% follows the settings in the original Z-LaVI paper~\cite{Yang:2022zlavi}.
\subsection{Experimental Details}
We adopt Google Image Search to retrieve relevant images and utilize the newly released Stable Diffusion~\cite{Rombach:2O22stablediffusion} for image synthesis.\footnote{\url{https://github.com/CompVis/stable-diffusion} and we use the \text{sd-v2-1} checkpoint.}
Following \citet{Yang:2022zlavi}, for each QA pair in \method, we obtain 100 images with each of the two methods. These 200 images are sorted using CLIP based on their similarity with the question. We preserve top-10 ($K=10$) images for each QA pair as the final image sources. The other experimental details, such as the model implementation, hyperparameters, and computing resources are presented in Appendix~\ref{app:exp}.

\subsection{Main Results}
\label{sec:main-results}
The main evaluation results of LLMs and VaLMs on \method are shown in Figure~\ref{fig:radar}. Here, we highlight several interesting findings. 

\textbf{Falcon and LLaMA excel in all four presented LLM model families, especially on the color and component sub-tasks.} As shown in Figure~\ref{fig:radar}(a, b), Falcon and LLaMA consistently outperform OPT and GPT across various subsets with both experimental settings, despite the shape subset. Particularly, LLaMA achieves a zero-shot accuracy of 41\% on the color subset, surpassing GPT-J with a considerable margin of 13\%; Falcon yields the highest few-shot accuracy of 76\% on the component subset, which amounts to 14\% absolution improvement over OPT. We further present the few-shot results of the largest available LLMs in their own model family in Figure~\ref{fig:radar}(c), where LLaMA-65B shows remarkable superiority over other counterparts.

\begin{figure}[t]
\centering
\includegraphics[width=0.75\columnwidth]{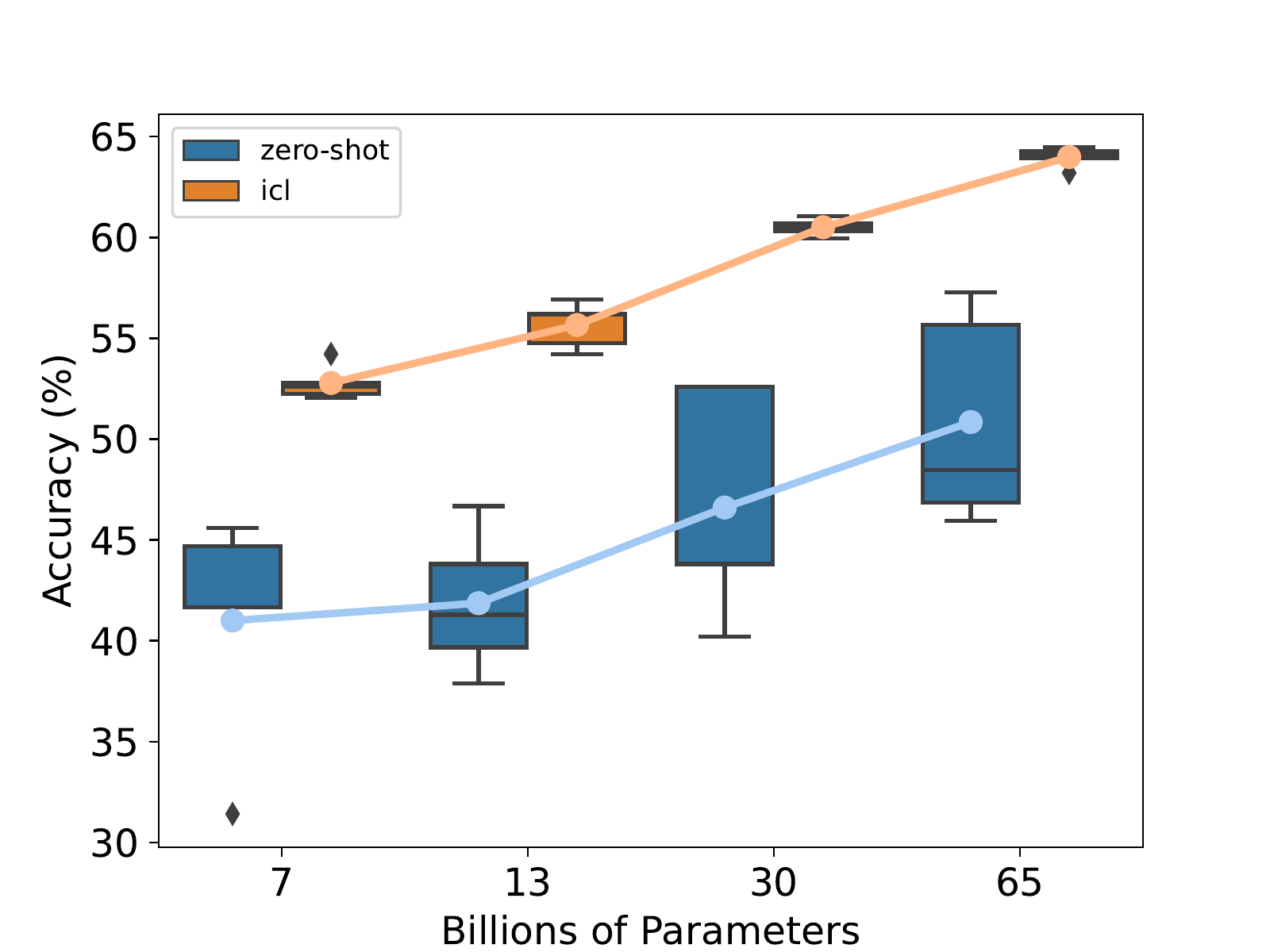}
\caption{Performance distribution of LLaMA on the color subset. Models with ICL achieve higher performance and show reduced variance across prompts.}
\label{fig:icl_box_plot}
\end{figure}

\textbf{In-context learning (ICL) not only improves the visual commonsense performance of LLMs but also reduces their variance across different prompts.} Comparing the results in Figure~\ref{fig:radar}(a) and \ref{fig:radar}(b), we found that given a few examples (i.e., with ICL), LLMs achieve consistent and remarkable improvement over themselves. For instance, LLaMA-7B with ICL achieves an average score of 62\% across five sub-tasks, with a 12\% improvement over the zero-shot result. We further show the performance distribution of LLaMA across different prompts in Figure~\ref{fig:icl_box_plot}, which illustrates that ICL not only improves the model's performance but also reduces its variance across different prompts. Further analysis is conducted in Section~\ref{sec:analysis}.

\textbf{VaLMs improve the visual commonsense ability of their LLM backbones, despite small performance gains on the shape subset.} 
As depicted in Figure~\ref{fig:radar}(d, e), BLIP-2 shows remarkable superiority over OPT on the color and material subset, with average accuracy improvements of 17\% and 15\%, respectively, which indicates that incorporating visual information indeed helps to improve LLMs' visual commonsense capabilities. However, the results also show that the performance gains of VaLMs are small on some sub-tasks: both BLIP-2 and MAGMA only achieve an 0.5\% accuracy improvement on the shape sub-task, while Z-LaVI even has performance drops. This demonstrates that VaLMs still have wide room for improvement. 

\textbf{ICL capability of VaLMs should be further valued.} As shown in Figure~\ref{fig:radar}(f), MAGMA with ICL achieves consistent improvements across all subsets over itself and the few-shot results of the \textit{frozen} LLM backbone, indicating that ICL could also improve VaLMs' visual commonsense performances. However, ICL has been somewhat under-investigated by previous VaLM research. For example, both Z-LaVI and BLIP-2 only support zero-shot evaluation, with the lack of ICL capability. We hope our research could draw more attention to future work on the ICL capability of VaLMs.

\section{Analysis}
\label{sec:analysis}
We further investigate the factors influencing the visual commonsense capabilities of LLMs and VaLMs. For instance, we find that a decent scale (e.g., 1.3B) could be a potential threshold for text-only LLMs to learn visual commonsense. We then analyze several influencing factors of VaLMs, such as image sources and image numbers.

\begin{figure}[t]
\centering
\includegraphics[width=0.8\columnwidth]{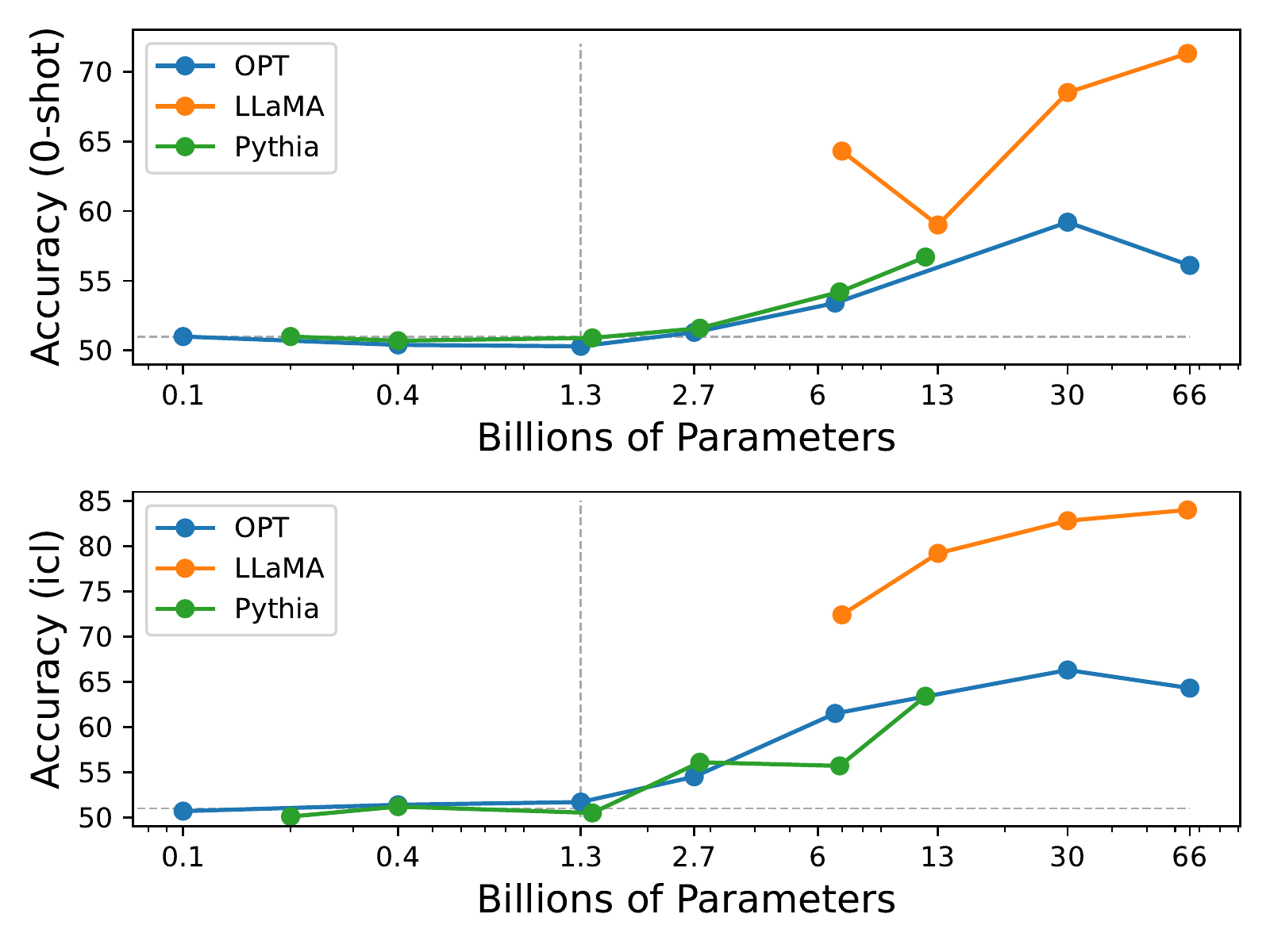}
\caption{Performance of LLMs on the component subset of \method. LLMs with a model size larger than 1.3B demonstrate emergent capabilities when solving the component sub-task.}
\label{fig:component_emergent}
\end{figure}

\begin{figure}[t]
\centering
\includegraphics[width=0.75\columnwidth]{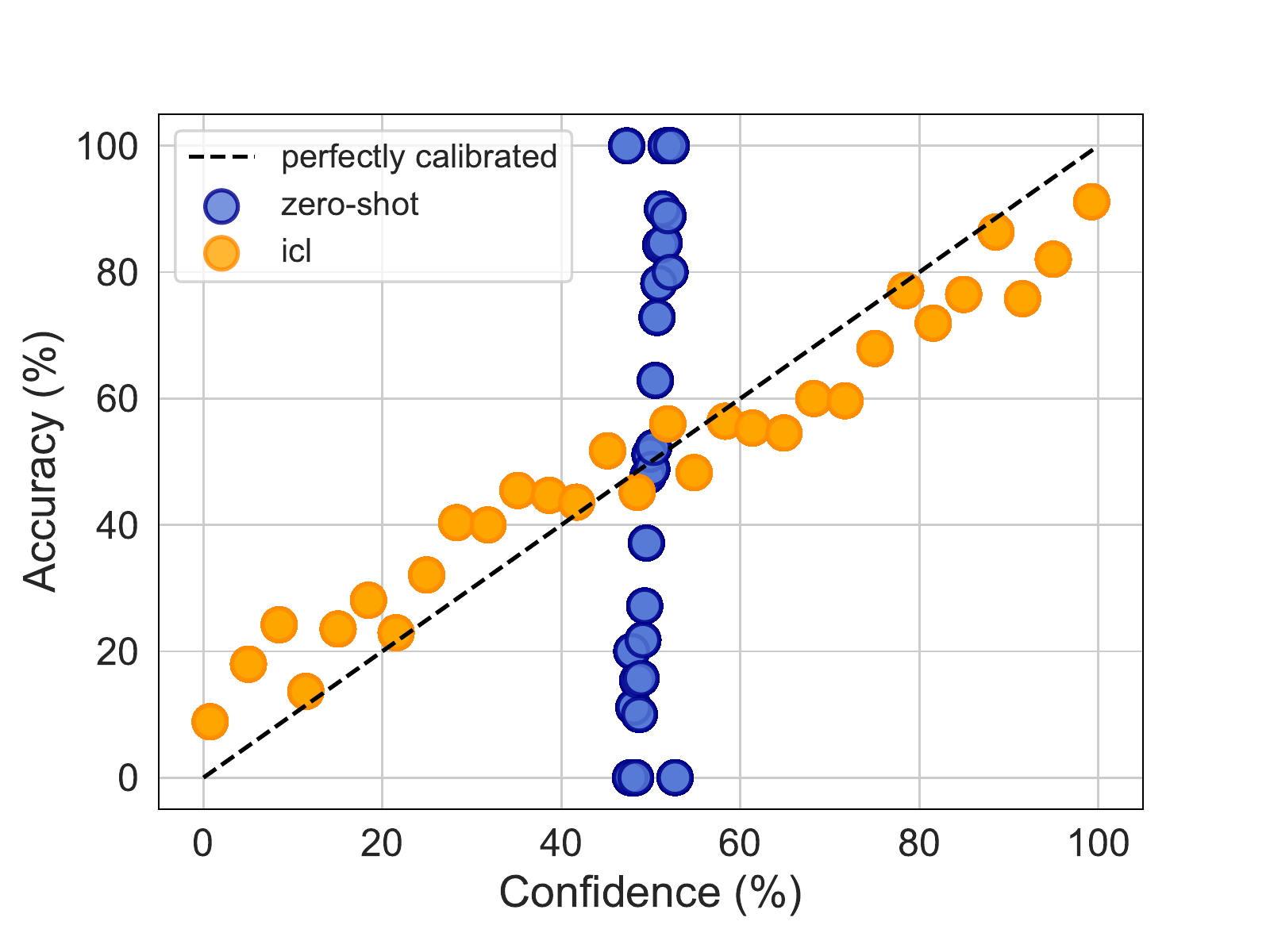}
\caption{Calibration results of LLaMA-7B on the component subset. ICL greatly enhances model calibration, significantly boosting the correlation between confidence and accuracy from $r=0.57$ to $r=0.98$.}
\label{fig:calibration}
\end{figure}

\paragraph{When (at what model scale) do text-only LLMs learn visual commonsense?}
We show the zero- and few-shot results of three LLM families on the component subset in Figure~\ref{fig:component_emergent}. Take the component sub-task as an example, we find that a decent scale (e.g., 1.3B) could be a starting point for LLMs to emerge with visual commonsense on the component\footnote{Please note that, as the evaluated LLMs (OPT and Pythia) both rely on the Pile~\cite{Gao:2023thepile} as their pre-training corpus, our findings may not be generalized to other LLMs.}: smaller models at sizes below 1.3B are unable to perform well on the task, with a performance close to random guessing (i.e., \textasciitilde50\% accuracy); while models with a size larger than 1.3B exhibit gradual performance improvements. For example, OPT-30B achieves 59\% and 66\% average accuracy with zero- and few-shot settings, respectively.

\paragraph{What is the effect of ICL on the calibration of LLMs?}
Ensuring the reliability of a model's predictions is a crucial aspect of model evaluation, as mis-calibrated models can lead to incorrect inferences and serious consequences in real-world applications~\cite{Guo:2017calibration}. To this end, we conducted a calibration analysis to evaluate whether the model confidence on visual commonsense reliably reflects the actual probability of the prediction being correct. Our analysis focuses on the calibration of LLaMA-7B on the component subset of \method. Results in Figure~\ref{fig:calibration} indicate that ICL significantly improves the calibration of the model, increasing the correlation between confidence and accuracy from $r=0.57$ to $r=0.98$. This is consistent with the aforementioned findings, suggesting that ICL improves the visual commonsense performance of LLMs.

\begin{table}[t]
\centering
\small
\scalebox{0.86}{
% \resizebox{\linewidth}{!}{
\begin{tabular}{c|ccccc|c}
\toprule
& \textbf{\textsc{Col.}} & \textbf{\textsc{Sha.}} & \textbf{\textsc{Mat.}} & \textbf{\textsc{Com.}} & \textbf{\textsc{Oth.}} & \textbf{\textsc{Avg}}\\
\midrule
OPT-2.7B & 25.8 & 39.9 & 40.2 & 51.3 & 48.1 & 41.1\\
\midrule
\ding{55} & 26.4 & 41.1 & 40.5 & 50.9 & 48.8 & 41.5 \\
\textsc{Random} & 20.1 & 38.9 & 42.0 & 51.9 & 47.4 & 40.9 \\
\textsc{Search} & \underline{44.2} & \textbf{40.5} & \underline{60.2} & 52.9 & 51.0 & \underline{49.8} \\
\textsc{Synthesis} & 43.2 & 39.5 & 59.2 & \underline{53.8} & \underline{51.3} & 49.4 \\
\textsc{Rank} & \textbf{44.7} & \underline{40.3} & \textbf{61.9} & \textbf{54.0} & \textbf{51.7} & \textbf{50.5} \\
\bottomrule
\end{tabular}}
\caption{Zero-shot results of BLIP-2 with OPT-2.7B on \method with various image sources. \ding{55} means no image is provided to the model.}
\label{tab:image-sources}
\end{table}

\paragraph{How do image sources influence VaLMs?}
Table~\ref{tab:image-sources} shows the ablation results of the image sources used in VaLMs.  
As illustrated, BLIP-2 with extra image sources (e.g., \textsc{Search}) brings large improvements. 
This supports \method's motivation, suggesting that previous visual commonsense evaluations undervalue VaLMs' potential, as VaLMs' visual commonsense demands relevant images as input to be suitably stimulated.
Among the various image sources, CLIP-ranked images yield the best performance, suggesting that aligning images closely with the question facilitates the generalization of related visual commonsense by models.
Thus, we use ranked images as the default image source for our main experiment and analysis. 

\begin{figure}[t]
\centering
\includegraphics[width=0.8\columnwidth]{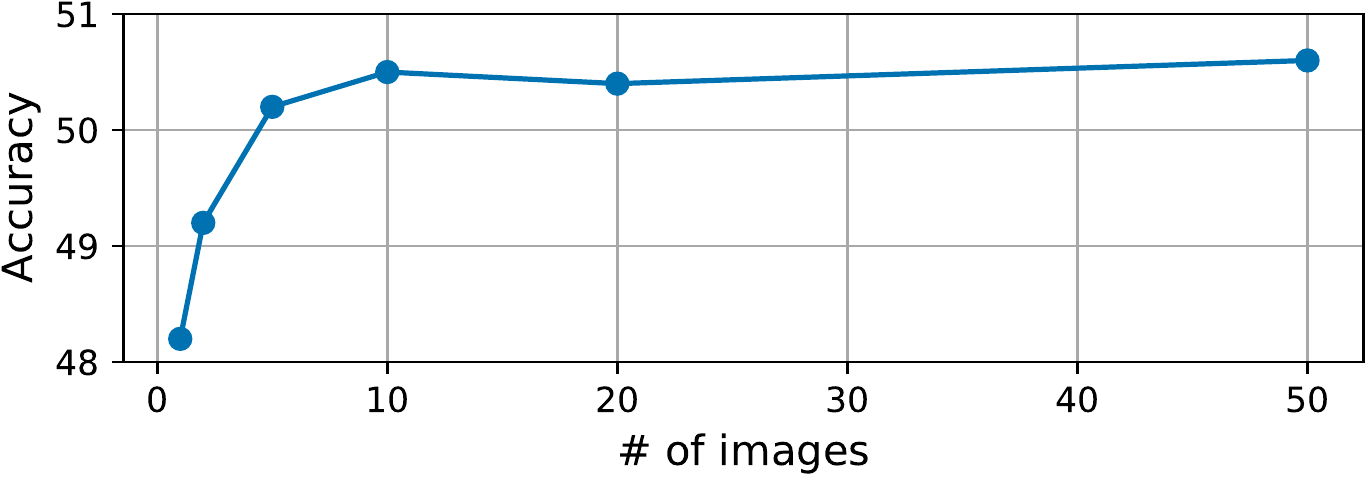}
\caption{Average performance on \method with various numbers of top-ranked images. The results are obtained with BLIP-2 (OPT-2.7B).}
\label{fig:image_nums}
\end{figure}

\begin{table}[t]
\centering
\small
\scalebox{0.86}{
% \resizebox{\linewidth}{!}{
\begin{tabular}{l|ccccc|c}
\toprule
\textbf{Models} & \textbf{\textsc{Col.}} & \textbf{\textsc{Sha.}} & \textbf{\textsc{Mat.}} & \textbf{\textsc{Com.}} & \textbf{\textsc{Oth.}} & \textbf{\textsc{Avg}}\\\midrule
12-in-1 & 70.0 & 65.6 & 65.0 & 72.7 & 67.9 & 68.2 \\
BLIP & \textbf{77.0} & 70.0 & \textbf{73.9} & 71.5 & 69.7 & 72.4 \\ \midrule
ChatGPT & 58.0 & 62.2 & 69.7 & 78.8 & 66.9 & 67.1 \\
$\llcorner$ \textit{w/} ICL & 67.9 & \textbf{75.0} & 72.3 & \textbf{87.5} & \textbf{76.2} & \textbf{75.8} \\ \midrule
Humans & 91.2 & 92.7 & 90.5 & 98.9 & 94.0 & 93.5 \\
\bottomrule
\end{tabular}}
\caption{Evaluation results of multimodal models and ChatGPT on \method. We report Top-1 accuracy results obtained by open-ended generation. We also show human performance in the last row as a reference.}
\label{tab:openend-others}
\end{table}

\paragraph{What is the typical number of images required to capture visual commonsense?}
We show the model's average performance on \method with various numbers of images in Figure~\ref{fig:image_nums}. The results show that the model's performance increases as the number of top-ranked images gradually increases from 1 to 10, indicating diverse image sources help the model to capture general visual commonsense. However, the improvement is marginal when the image number is larger than 10. 

\section{Visual Commonsense in Other Models}
It is worth noting that, as a general visual commonsense dataset, \method supports various types of models and evaluation settings. Except for the evaluation setting in our main results, we also evaluate several models in the setting of open-ended generation. Specifically, we select two widely-used multimodal models, 12-in-1~\cite{Lu:202012in1} and BLIP~\cite{Li:2022blip} that are finetuned on the VQAv2 dataset~\cite{Goyal:2017VQAv2} and the famous RLHF model, ChatGPT (\texttt{gpt-3.5-turbo}) for evaluation.\footnote{The evaluation details are illustrated in Appendix~\ref{sec:other-eva-details}.} As illustrated in Table~\ref{tab:openend-others}, the multimodal models finetuned on VQAv2 show strong performance on \method, especially on the color sub-task, with relatively small model scales (e.g., 583M of BLIP). ChatGPT with ICL achieves the best average accuracy score of 75.8\% across all compared models. However, it still has a considerable performance gap with humans, which has an average performance of 93.5\%.

\section{Conclusion}
In this paper, we introduced \method, a comprehensive human-annotated dataset for evaluating visual commonsense using both textual and visual inputs. We conducted extensive experiments to evaluate the visual commonsense of both unimodal LLMs and VaLMs using \method. Our results demonstrate the varying degrees of visual commonsense knowledge present in different models, as well as the factors that contribute to the acquisition and enhancement of this knowledge. Additionally, we offer insights into the emergent abilities of LLMs and the strengths of VaLMs in the realm of visual commonsense.

\section*{Limitations}
While our study provides valuable resources and insights into the visual commonsense knowledge of LLMs and VaLMs, several limitations need to be acknowledged. Firstly, due to the high cost of region-based human annotation, the \method dataset only covers 1,000 ImageNet categories, which may not cover all real-world scenarios. Therefore, it is possible that models may perform differently on other types of images that are outside of the \method categories.

Additionally, our study is limited to zero- and few-shot visual commonsense evaluation and only considers models that have been pretrained on large amounts of text data. Thus, it remains unclear whether fine-tuning on visual commonsense tasks would improve the performance of LLMs and VaLMs. Therefore, this may not fully capture a model's ability to understand visual commonsense in real-world scenarios where prior knowledge may be available.
Besides, although we explored the factors that affect the visual commonsense knowledge of large models, it is still challenging to interpret how the models acquire this knowledge. 

Overall, our study provides a foundation for future work in the field of visual commonsense knowledge and its applications. However, additional research is necessary to address the aforementioned limitations and further advance the field.

\section*{Acknowledgements}
This paper is supported by the National Key Research and Development Program of China 2020AAA0106700 and NSFC project U19A2065.

\bibliography{anthology,custom}
\bibliographystyle{acl_natbib}

\clearpage

\appendix

\section*{Appendix}
\section{Annotation Details}
\label{sec:ui}
\label{appendix:annotation}
In this section, we will provide a comprehensive overview of our annotation process, including the guidelines we follow, the user interface we use, the hierarchical supervision process we employ to ensure data quality, and our payment policy.
\subsection{Annotation Guidelines}
The annotation of \method involves observing 20-50 images of a given category, finding a vision feature of the category, checking if it conforms to most of the images and our commonsense of life, and then writing a simple question-answer (QA) about this vision feature. The QA should contain one question and one correct answer.

The vision features can be object-based (such as the color, shape, material, and spotted/striped patterns of the whole object) or region-based (such as the color, shape, and material of a certain part of the object). They are features that can be seen through the images.

The annotation pipeline involves looking at the 20-50 images given and finding a common vision feature of the category. For example, "The shape of the dorsal fin of the tiger shark is triangle". The annotators check if this feature conforms to their commonsense of life and if it is written. Then, one QA is created, such as "What is the shape of the dorsal fin of the tiger shark? Triangle".

The following rules must be followed during the annotation process:
\begin{itemize}
    \item The question should contain the name of the category. Otherwise, the submission will not be passed.
    \item If the annotator cannot think of a question that can be written or the images cannot be displayed, the annotator can skip this category.
    \item The first letter of the question needs to be capitalized.
    \item The end of the question needs to be a question mark.
    \item Please do not write lots of Yes/No questions. These questions are more likely to be rejected.
    We encourage the annotators to write more diverse answers.
\end{itemize}

In the annotation examples, we describe how the correct QA is created. Annotators can write their own QA according to this pipeline. The rejected examples include cases where the QA has been written before, vision features cannot be found in the images, or the QA is not about vision features. Additionally, the QA should conform to our commonsense of life, be strongly related to the category, and be about a specific feature of the category.

In summary, the annotation guidelines of \method involve observing images of a category, finding a vision feature, and creating a QA about the feature that conforms to our commonsense of life and follows the rules outlined above. These guidelines ensure the accuracy and consistency of the annotations, making the dataset suitable for use in various applications

\begin{table*}[t]
\centering
\small
\begin{tabular}{ll}
\toprule
\textbf{Subset} &\textbf{Answer Candidates}\\
\midrule
Color & brown (tan), black, white, yellow (gold, golden), green, gray (grey), red, orange, blue, silver, pink \\ \midrule
\multirow{2}{*}{Shape} 
& round (circle), rectangle (rectangular), triangle (triangular), square, oval, curved, cylinder, straight\\
& cone, curly, heart, star \\ \midrule
\multirow{2}{*}{Material}
& metal (steel, iron), wood (wooden), plastic, cotton (yarn, wool), glass, fabric (nylon, silk, rope) \\
& stone (rock), rubber, ceramic (porcelain), cloth (denim), leather, flour (dough, bread), paper, clay, wax \\ 
& concrete \\ \midrule
Component & yes, no \\ \midrule
Others (yes/no) & yes, no \\ \midrule
Others (numbers) & 2 (two), 4 (four), 6 (six), 1 (one), 8 (eight), 3 (three), 5 (five)  \\ \midrule
\multirow{4}{*}{Others (others)}
& long, small, short, large, forest (jungle, woods), water, ocean, big, tree (branch), ground, tall, wild \\
& outside, thin, head, thick, circle, brown, soft, land, neck, rough, chest, smooth, fur, hard, top, plants \\
& black, metal, books, vertical, lake (pond), grass, road, sky, front, kitchen, feathers, stripes, baby, hair \\
& feet, mouth, female, table \\
\bottomrule
\end{tabular}
\caption{The answer set of all subsets in \method. Inside the parentheses are attributes grouped into the same answer candidate.}
\label{tab:answer-set}
\end{table*}

\subsection{Annotation UI}
\begin{figure*}[t]
\centering
\includegraphics[width=0.9\textwidth]{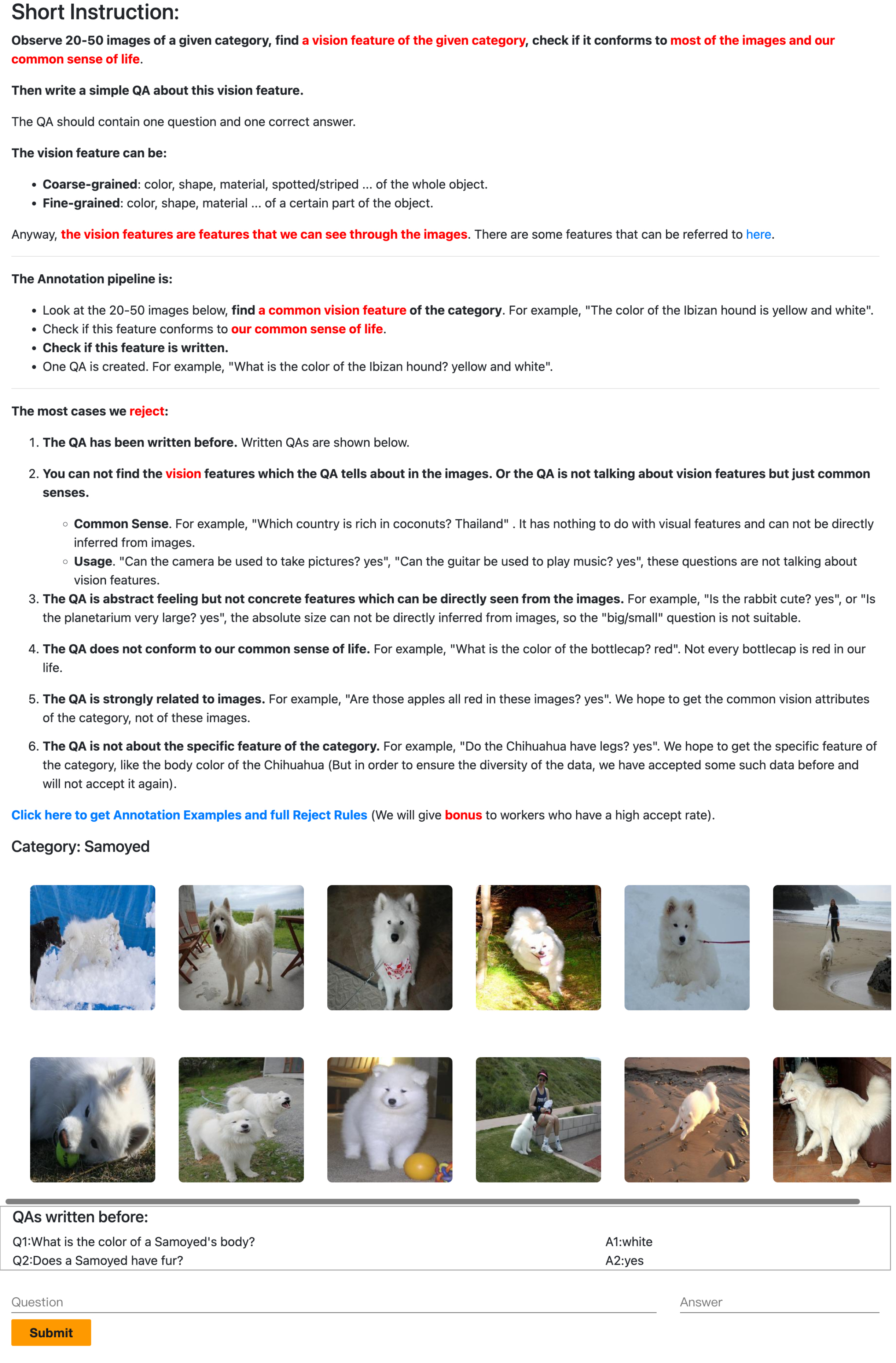}
\caption{Screenshot of the \method annotation UI, featuring task instructions, the annotation pipeline, and the most common reasons for rejecting prompts during the annotation process. The interface displays 20-50 images of a given category, and the task instruction guides the annotator to identify a common vision feature and create a simple QA about it. The annotation pipeline includes checking the conformity to commonsense and avoiding pre-written QAs. Annotations are focused on visual features, not commonsense or non-visual attributes.}
\label{fig:ui}
\end{figure*}

Figure~\ref{fig:ui} shows the annotation user interface used in our human annotation process for model knowledge assessment. The interface consists of three main parts: the task instruction, the annotation pipeline, and the most common cases we reject. The task instruction provides clear guidance for the annotators on how to write effective prompts to assess the model knowledge. The annotation pipeline displays the generated text by the model and allows the annotators to refine their prompts until the generated text matches the expected target answer. The rejected cases section provides examples of prompts that do not meet the criteria and serves as a reference for the annotators to avoid such mistakes. The user interface design is intuitive and user-friendly, which greatly improves the efficiency and accuracy of the human annotation process.

\subsection{Hierarchical Supervision}
\label{sec:annotation-supervision}
To ensure high-quality annotation, we have implemented a hierarchical supervision process. During the annotation phase, examiners cross-check the annotation results, and annotators who are excessively rejected receive a warning. Those who exceed the warning limit are removed. Additionally, during the examination phase, a random sample check of the examination results is performed by five authors, and examiners with low-quality checks also receive a warning. The hierarchical supervision process guarantees the high-quality execution of the entire annotation process. 

\subsection{Payment Policy}
\label{pricing-strategy}
We compensated the crowd workers with varying rates based on the workload and quality of their work. In Phase 1, workers were tasked with summarizing shared visual characteristics from 50 images and creating QA pairs. If their work was accepted in Phase 2, they would receive \$0.50 for each sample. However, if their work was rejected, they would only earn \$0.10 per sample. In Phase 2, annotators were paid \$0.30 per sample for cross-checking tasks. On average, the annotators received approximately double the local minimum wage per hour.

\section{Details of \method}
\label{sec:details-of-imagenetvc}

\subsection{Details of the Others subset}
\begin{figure}[htbp]
\centering
\includegraphics[width=0.8\columnwidth]{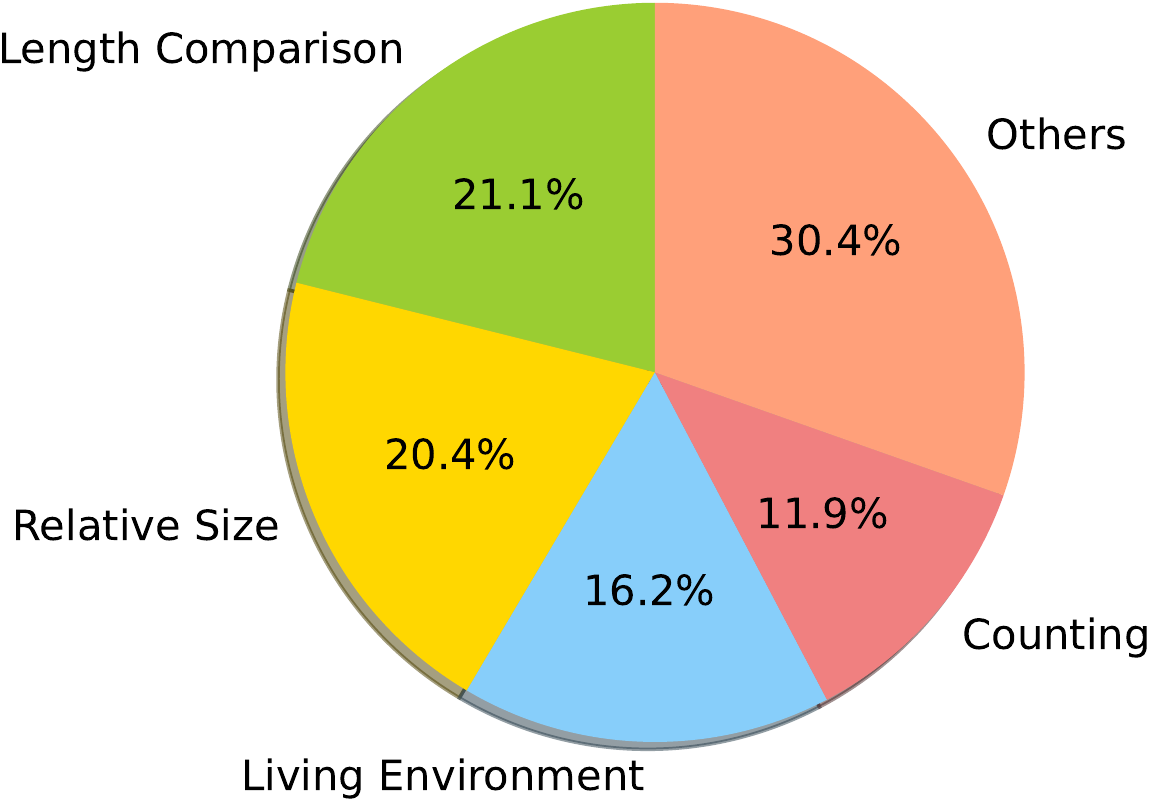}
\caption{Detailed composition of the Others subset.}
\label{fig:other-composition}
\end{figure}

Annotated QA pairs that do not belong to the four specified sub-tasks (e.g., color, shape, material, and component) will be categorized into the Others subset. Therefore, The Others subset contains a more diverse range of QA samples, which is more challenging. Figure~\ref{fig:other-composition} illustrates the detailed composition of QA types in the Others subset, which covers various topics such as length comparison (21\%), relative size (20\%), living environment (16\%), counting (12\%), etc.

\subsection{Answer Set}
Considering that the results of open-ended generation are uncontrollable, we evaluate all models with constrained decoding in our main experiments. Table~\ref{tab:answer-set} shows the list of all possible answers in \method. Besides, we noticed that LLMs tend to predict ``yes/no'' or numerical answers when evaluated on the others subset. Thus, we split the others subset into three small test sets, containing answer types of ``yes/no'', numbers, and other answers, respectively.

\subsection{More Qualitative Examples}
We present additional qualitative examples in Table~\ref{tab:qual_examples}, which compare the predictions made by various models. The comparisons between OPT-7B and BLIP-2 demonstrate the effectiveness of incorporating visual information in enhancing the visual commonsense capabilities of LLMs. However, these leading models, including ChatGPT, also encounter difficulties in certain challenging cases, such as determining the color of a flamingo's beak tip. Besides, these examples highlight ChatGPT's tendency to prioritize selecting the most commonly associated property of an object as the answer, rather than considering the properties of the specific region in question. We hypothesize that this behavior may be attributed to the higher frequency of these common attributes co-occurring with the object in the pre-training text corpus.

\begin{table*}[t]
\centering
\small
\begin{tabular}{p{2.4cm}p{10.2cm}}
\toprule
\textbf{Related Images} &
    \begin{minipage}{.6\textwidth}
        \includegraphics[height=15mm]{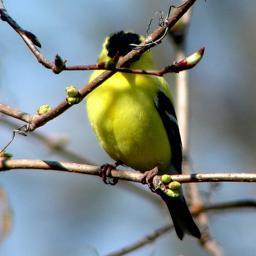}
        \includegraphics[height=15mm]{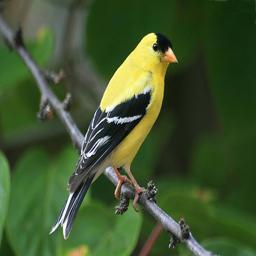}
        \includegraphics[height=15mm]{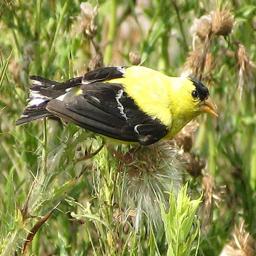}
        \includegraphics[height=15mm]{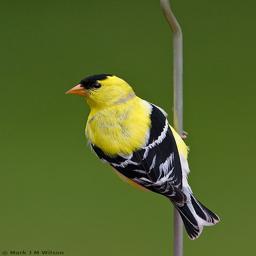}
        \includegraphics[height=15mm]{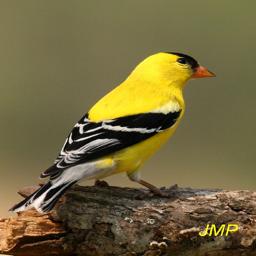}
        \includegraphics[height=15mm]{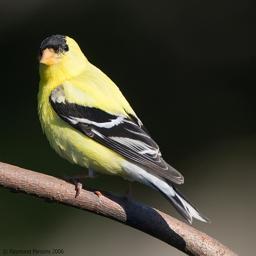}
        \vspace{2pt}
    \end{minipage}
    ...
\\  
\textbf{Question} & \textbf{\textit{What color is a goldfinch's tail?}} \\  
\textbf{Answer} & OPT-7B: red \redno, BLIP-2: black \greenyes, ChatGPT: yellow \redno.\\ \midrule  
\textbf{Related Images} &
    \begin{minipage}{.6\textwidth}
        \includegraphics[height=15mm]{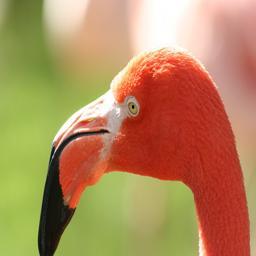}
        \includegraphics[height=15mm]{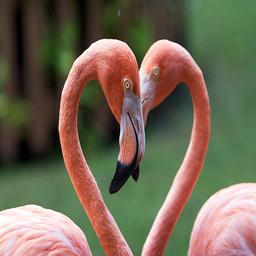}
        \includegraphics[height=15mm]{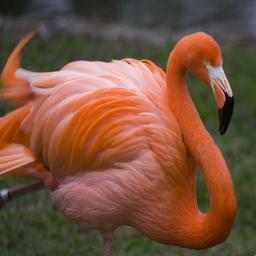}
        \includegraphics[height=15mm]{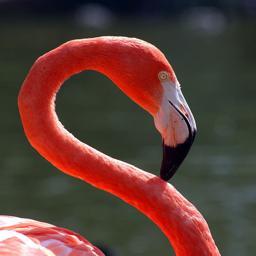}
        \includegraphics[height=15mm]{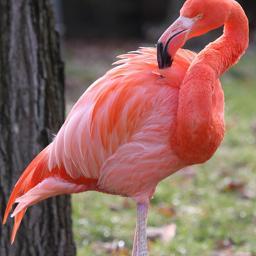}
        \includegraphics[height=15mm]{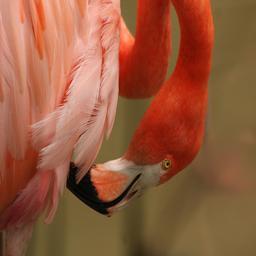}
        \vspace{2pt}
    \end{minipage}
    ...
\\  
\textbf{Question} & \textbf{\textit{What is the color of the flamingo's beak tip?}} \\   
\textbf{Answer} & OPT-7B: green \redno, BLIP-2: red \redno, ChatGPT: pink \redno, (Ground Truth: black). \\ \midrule  
\textbf{Related Images} &
    \begin{minipage}{.6\textwidth}
        \includegraphics[height=15mm]{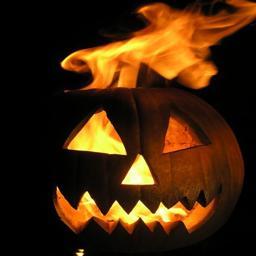}
        \includegraphics[height=15mm]{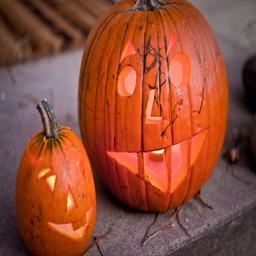}
        \includegraphics[height=15mm]{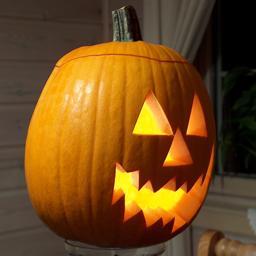}
        \includegraphics[height=15mm]{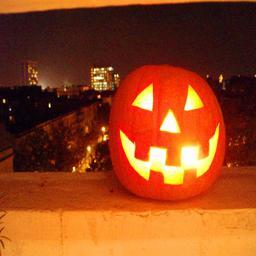}
        \includegraphics[height=15mm]{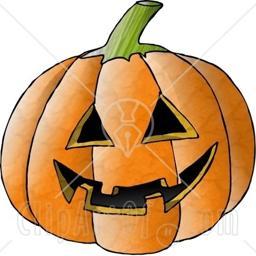}
        \includegraphics[height=15mm]{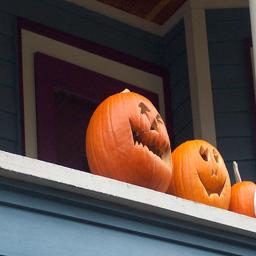}
        \vspace{2pt}
    \end{minipage}
    ...
\\  
\textbf{Question} & \textbf{\textit{What is the shape of the eyes of the jack-o'-lantern?}} \\ 
\textbf{Answer} & OPT-7B: cylinder \redno, BLIP-2: square \redno, ChatGPT: triangle \greenyes. \\ \midrule  
\textbf{Related Images} &
    \begin{minipage}{.6\textwidth}
        \includegraphics[height=15mm]{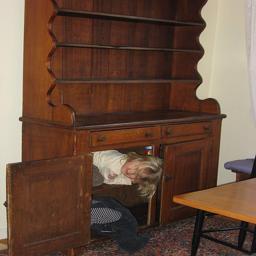}
        \includegraphics[height=15mm]{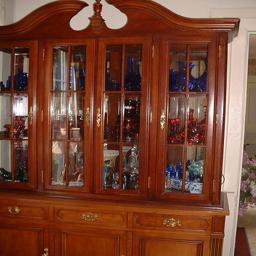}
        \includegraphics[height=15mm]{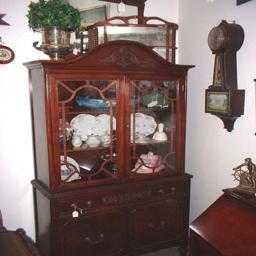}
        \includegraphics[height=15mm]{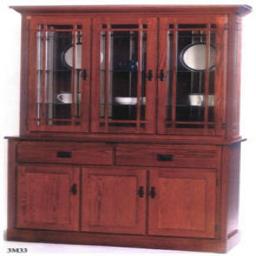}
        \includegraphics[height=15mm]{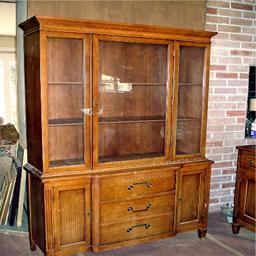}
        \includegraphics[height=15mm]{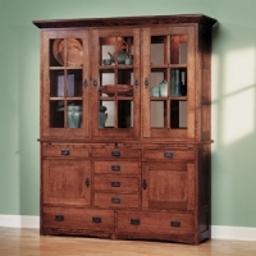}
        \vspace{2pt}
    \end{minipage}
    ...
\\  
\textbf{Question} & \textbf{\textit{What's the material of a china cabinet?}} \\ 
\textbf{Answer} & OPT-7B: cotton \redno, BLIP-2: wood \greenyes, ChatGPT: ceramic \redno. \\ \midrule  
\textbf{Related Images} &
    \begin{minipage}{.6\textwidth}
        \includegraphics[height=15mm]{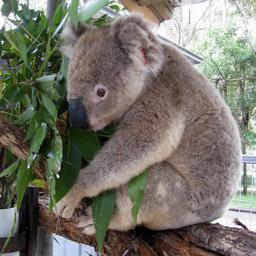}
        \includegraphics[height=15mm]{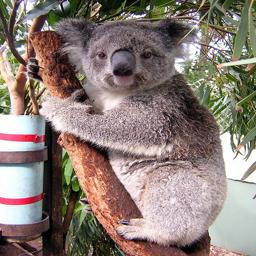}
        \includegraphics[height=15mm]{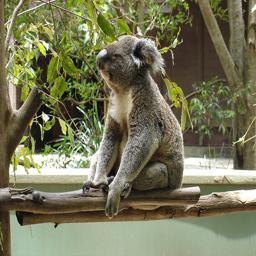}
        \includegraphics[height=15mm]{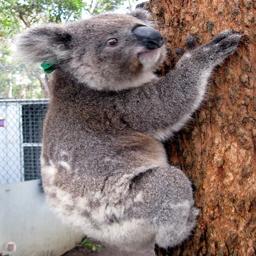}
        \includegraphics[height=15mm]{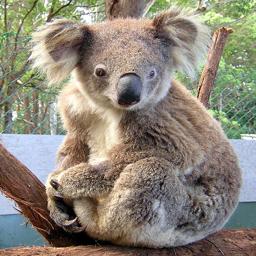}
        \includegraphics[height=15mm]{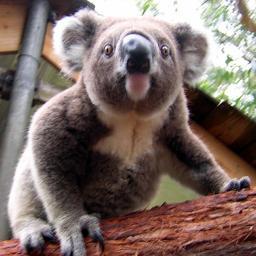}
        \vspace{2pt}
    \end{minipage}
    ...
\\  
\textbf{Question} & \textbf{\textit{Does the koala have a tail? }} \\   
\textbf{Answer} & OPT-7B: yes \redno, BLIP-2: no \greenyes, ChatGPT: yes \redno. \\

\bottomrule
\end{tabular}
\caption{
    More qualitative examples comparing different models on \method. Here we demonstrate comparisons with OPT-7B, BLIP-2, and ChatGPT.
}
\label{tab:qual_examples}
\end{table*}

\begin{table}[t]
\centering
\small
\begin{tabular}{l}
\toprule
\textbf{Prompts} \\
\midrule
{[}QUESTION{]} {[}ANSWER{]}. \\
{[}QUESTION{]} Answer: {[}ANSWER{]}.  \\
{[}QUESTION{]} The answer is {[}ANSWER{]}.  \\
Question: {[}QUESTION{]} Answer: {[}ANSWER{]}. \\
Question: {[}QUESTION{]} The answer is {[}ANSWER{]}. \\
\bottomrule
\end{tabular}
\caption{The prompts we utilize for LLM and VaLM evaluation on \method.}
\label{tab:prompts}
\end{table}

\section{Experimental Details}
\subsection{Multiple Prompts}
\label{sec:eva-prompts}
Table~\ref{tab:prompts} shows all the intuitive prompt templates utilized to evaluate the LLMs and VaLMs in our main experiments. We do not tune the prompt for each subset in \method.

In our data quality experiments, we utilize the original prompts provided by \citet{Zhang:2022vicomte} for ViComTe evaluation while adopting the prompts in Table~\ref{tab:prompts} for \method.

\begin{table*}[t]
\centering
\small
\setlength{\tabcolsep}{3pt}
%\resizebox{1.0\linewidth}{!}{
\begin{tabular}{@{}cl|c|cc|cc|cc|cc|cc|c@{}}
\toprule
\multicolumn{2}{c|}{\multirow{2}{*}{\textbf{Models}}} & \multirow{2}{*}{\textbf{\#Param}} & \multicolumn{2}{c|}{\textbf{\textsc{Color}}} & \multicolumn{2}{c|}{\textbf{\textsc{Shape}}} & \multicolumn{2}{c|}{\textbf{\textsc{Mater.}}} & \multicolumn{2}{c|}{\textbf{\textsc{Compo.}}} & \multicolumn{2}{c|}{\textbf{\textsc{Others}}} & \multirow{2}{*}{\textbf{\textsc{Avg}}} \\ \cmidrule(lr){4-13}
\multicolumn{2}{c|}{} &  & 0-shot & 5-shot & 0-shot & 5-shot & 0-shot & 5-shot & 0-shot & 5-shot & 0-shot & 5-shot & \\
\midrule
&Random$^*$ & - & 7.7 & 7.7 & 9.0 & 9.0 & 6.1 & 6.1 & 49.8 & 49.8 & 24.3 & 24.3 & 19.4 \\
\midrule
\multirow{7}{*}{OPT}
&OPT-125m & 125M & 16.3 & 7.6 & 18.7 & 12.4 & 25.2 & 24.8 & 51.0 & 50.7 & 38.4 & 39.4 & 28.5  \\
&OPT-350m & 350M & 19.9 & 23.1 & 18.4 & 25.8 & 27.7 & 26.0 & 50.4 & 51.4 & 38.2 & 42.3 & 32.3 \\
&OPT-1.3B & 1.3B & 26.5 & 27.1 & 32.7 & 42.6 & 37.3 & 41.3 & 50.3 & 51.7 & 44.2 & 53.0 & 40.7 \\
&OPT-2.7B & 2.7B & 25.8 & 33.4 & 39.9 & 49.7 & 40.2 & 58.7 & 51.3 & 54.5 & 48.1 & 55.6 & 45.8 \\
&OPT-6.7B & 6.7B & 31.1 & 37.1 & 38.6 & 51.2 & 46.1 & 61.4 & 53.4 & 61.5 & 51.9 & 52.9 & 48.5 \\
&OPT-30B & 30B & 31.7 & 41.1 & 40.5 & 59.2 & 49.8 & 70.4 & 59.2 & 66.3 & 54.4 & 59.7 & 53.2 \\
&OPT-66B & 66B & 34.1 & 44.0 & 40.6 & 57.1 & 47.3 & 68.4 & 56.1 & 64.3 & 55.5 & 60.8 & 52.8 \\
\midrule
\multirow{3}{*}{GPT}
&GPT-Neo-1.3B & 1.3B & 23.6 & 24.3 & 34.1 & 47.9 & 35.9 & 40.2 & 51.8 & 51.4 & 45.0 & 53.0 & 40.8 \\
&GPT-Neo-2.7B & 2.7B & 24.1 & 31.1 & 25.1 & 46.5 & 39.3 & 47.1 & 52.0 & 51.0 & 46.6 & 53.0 & 41.6 \\
&GPT-J-6B & 6B & 27.9 & 39.4 & 39.8 & 56.3 & 49.2 & 63.8 & 52.4 & 59.0 & 51.9 & 58.0 & 49.8 \\
\midrule
\multirow{6}{*}{Pythia}
&Pythia-160m & 160M & 18.6 & 23.0 & 24.7 & 34.8 & 24.3 & 32.0 & 51.0 & 50.1 & 39.8 & 42.2 & 34.1 \\
&Pythia-410m & 410M & 11.2 & 25.8 & 20.9 & 39.1 & 36.8 & 44.2 & 50.7 & 51.2 & 43.8 & 47.4 & 37.1 \\
&Pythia-1.4B & 1.4B & 18.8 & 29.8 & 36.5 & 41.5 & 39.0 & 55.6 & 50.9 & 50.5 & 47.3 & 52.4 & 42.2 \\
&Pythia-2.8B & 2.8B & 18.6 & 34.0 & 31.6 & 53.5 & 43.2 & 46.1 & 52.6 & 56.1 & 49.6 & 55.7 & 44.1 \\
&Pythia-6.9B & 6.9B & 24.7 & 38.4 & 30.7 & 52.6 & 42.2 & 60.8 & 54.2 & 55.7 & 51.9 & 56.5 & 46.8 \\
&Pythia-12B & 12B & 27.6 & 43.3 & 34.6 & 47.7 & 46.3 & 63.8 & 56.7 & 63.4 & 54.2 & 62.4 & 50.0 \\
\midrule
\multirow{2}{*}{Falcon}
&Falcon-7B & 7B & 36.9 & 50.3 & 35.8 & 60.4 & 55.8 & 69.8 & 59.3 & 75.5 & 54.9 & 58.4 & 55.7 \\
&Falcon-40B & 40B & 46.4 & 56.2 & 42.2 & 62.5 & 57.5 & 71.9 & 69.0 & 79.6 & 59.9 & 68.9 & 61.7 \\
\midrule
\multirow{4}{*}{LLaMA}
&LLaMA-7B & 7B & 41.0 & 52.8 & 34.9 & 54.4 & 50.9 & 69.9 & 64.3 & 72.4 & 57.0 & 58.9 & 55.7 \\
&LLaMA-13B & 13B & 41.9 & 55.7 & 40.8 & 65.9 & 52.0 & \textbf{72.1} & 59.0 & 79.2 & 58.5 & 64.2 & 58.9 \\
&LLaMA-30B & 30B & 46.6 & 60.5 & 41.5 & 52.7 & \textbf{58.2} & 72.0 & 68.5 & 82.8 & 59.4 & 70.3 & 61.3 \\
&LLaMA-65B & 65B & \textbf{50.8} & \textbf{64.0} & \textbf{46.7} & \textbf{70.4} & 57.5 & 71.8 & \textbf{71.3} & \textbf{84.0} & \textbf{61.1} & \textbf{73.0} & \textbf{65.1} \\
\bottomrule
\end{tabular}%}
\caption{Evaluation results of large language models (LLMs) in \method. We report the mean Top-1 accuracy (\%) over 5 different prompts. $^*$: We report random results over 5 different runs for comparison. We show the best results in \textbf{boldface}.}
\label{tab:scaling-results}
\end{table*}

\subsection{Details of Human Assessment}
\label{app:human-assess}
In the human assessment process of ImageNetVC, annotators were presented with pairs of mutually exclusive random 32 instances from both ImageNetVC and other datasets (e.g., ViComTe) each time. In the whole process of comparison evaluation, we conducted multiple rounds of human assessment based on the dataset containing fewer test samples. Specifically, there are 556 comparison pairs used for evaluation between ImageNetVC and ViComTe, and 510 pairs used for evaluation between ImageNetVC and ChatGPT generated data.  

Annotators were instructed to evaluate these instances based on the overall quality of the data, choosing the better side when considering three factors respectively: diversity, difficulty, and factuality. The final results were computed and demonstrated as percentages in Figure~\ref{fig:quality_eval}. For instance, the Diversity score depicted in Figure~\ref{fig:quality_eval}(a) signifies that in 86\% of cases, the evaluators found ImageNetVC samples to either outperform or match those from ViComTe concerning diversity.

\subsection{Model Implementation}
\label{app:exp}
All the LLMs are implemented based on the Huggingface API\footnote{\url{https://huggingface.co}}. For VaLMs, we utilize the official release of Z-LAVI\footnote{\url{https://github.com/YueYANG1996/Z-LaVI}}, MAGMA\footnote{\url{https://github.com/Aleph-Alpha/magma}}, and BLIP-2\footnote{\url{https://github.com/salesforce/LAVIS/tree/main/projects/blip2}} for evaluation. The CLIP model is adapted from the OpenAI’s public source\footnote{\url{https://github.com/openai/CLIP}}. We utilize the vision-language pretrained checkpoints of VaLMs instead of task-specific finetuned models to evaluate their zero- and few-shot capabilities. The details of VaLMs, including the visual encoder architecture and pretraining data, are included in Appendix~\ref{sec:valm-intro}. All hyperparameters of VaLMs are the same as that of their origin paper. We conduct our experiments on 6 NVIDIA A40 GPUs with 48GB memory.

\begin{table}[t]
\centering
\small
\begin{tabular}{l|ccc}
\toprule
\textbf{Models} & \begin{tabular}[c]{@{}c@{}}\#Extra \\Params\end{tabular} & \begin{tabular}[c]{@{}c@{}}Visual \\Encoder\end{tabular} & \begin{tabular}[c]{@{}c@{}}\#Pretrained \\Images\end{tabular}\\
\midrule
Z-LaVI  &150M  & ViT-B/32 &- \\
MAGMA   &400M  & RN50x16 & 25M\\
BLIP-2  &1.1B  & ViT-G/14 & 129M  \\
\bottomrule
\end{tabular}
\caption{Details of VaLMs evaluated on \method.}
\label{tab:valm-details}
\end{table}

\begin{table*}[t]
\centering
\small
\resizebox{0.9\linewidth}{!}{
\begin{tabular}{cl|c|llllll}
\toprule
\multicolumn{2}{c|}{\textbf{Models}} & \textbf{\#Param.} & \textbf{\textsc{Color}} & \textbf{\textsc{Shape}} & \textbf{\textsc{Mater.}} & \textbf{\textsc{Compo.}} & \textbf{\textsc{Others}} & \textbf{\textsc{Avg}}\\
\midrule
&Random & - & 7.7 & 9.0 & 6.1 & 49.8 & 24.3 & 19.4 \\
\midrule
\multirow{3}{*}{\begin{tabular}[c]{@{}c@{}}VaLM \\(2.7B)\end{tabular}}
&OPT-2.7B & 2.7B & 25.8 & \underline{39.9} & 40.2 & 51.3 & \underline{48.1} & 41.1\\
&$\llcorner$ \textit{w/} Z-LaVI & 2.9B & \underline{37.3} \ua{11.5} & 31.2 \da{8.7} & \underline{46.6} \ua{6.4} & \underline{51.5} \ua{0.2} & 46.4 \da{1.7} & \underline{42.6} \ua{1.5} \\
&$\llcorner$ \textit{w/} BLIP-2 & 3.8B & \textbf{44.7} \ua{18.9} & \textbf{40.3} \ua{0.4} & \textbf{61.9} \ua{21.7} & \textbf{54.0} \ua{2.7} & \textbf{51.7} \ua{3.6} & \textbf{50.5} \ua{9.4} \\
\midrule
\multirow{3}{*}{\begin{tabular}[c]{@{}c@{}}VaLM \\(6B)\end{tabular}}
&GPT-J-6B & 6B & 27.9 & \underline{39.8} & 49.2 & 52.4 & \underline{51.9} & 44.2 \\
&$\llcorner$ \textit{w/} Z-LaVI & 6.2B & \underline{38.2} \ua{10.3} & 32.7 \da{7.1} & \underline{51.9} \ua{2.7} & \underline{52.5} \ua{0.1} & 48.6 \da{3.3} & \underline{44.8} \ua{0.6} \\
&$\llcorner$ \textit{w/} MAGMA & 6.4B & \textbf{42.9} \ua{15.0} & \textbf{40.3} \ua{0.5} & \textbf{60.9} \ua{11.7} & \textbf{54.6} \ua{2.2} & \textbf{53.9} \ua{2.0} & \textbf{50.5} \ua{6.3} \\
\midrule
\multirow{3}{*}{\begin{tabular}[c]{@{}c@{}}VaLM \\(6.7B)\end{tabular}}
&OPT-6.7B & 6.7B & 31.1 & \underline{38.6} & 46.1 & \underline{53.4} & \underline{51.9} & 44.2 \\
&$\llcorner$ \textit{w/} Z-LaVI & 6.9B & \underline{38.8} \ua{7.7} & 33.2 \da{5.4} & \underline{49.5} \ua{3.4} & 52.2 \da{1.2} & 49.6 \da{2.3} & \underline{44.7} \ua{0.5} \\
&$\llcorner$ \textit{w/} BLIP-2 & 7.8B & \textbf{47.9} \ua{16.8} & \textbf{39.1} \ua{0.5} & \textbf{61.3} \ua{15.2} & \textbf{58.4} \ua{5.0}  & \textbf{57.4} \ua{5.5} & \textbf{52.8} \ua{8.6} \\
\bottomrule
\end{tabular}}
\caption{Zero-shot probing results of visually-augmented language models (VaLMs) in \method. We report the mean accuracy (\%) results in 5 different prompts. Numbers that are highlighted in \hlsecondarytab{orange} represent the percentage of improvement and \hlprimarytab{blue} denotes the percentage of performance drop.}
\label{tab:valm-results}
\end{table*}

\subsection{Model Details}
\label{sec:valm-intro}
We mainly conduct our evaluations with the following VaLMs in our experiments:
\begin{itemize}
    \item \textbf{Z-LaVI}~\cite{Yang:2022zlavi} introduces a zero-shot framework that ensembles the solutions of LLMs and CLIP~\cite{Radford:2021clip} to handle plain language tasks. The extra visual inputs of CLIP are obtained with web search and synthesis methods.
    \item  \textbf{MAGMA}~\cite{Eichenberg:2022magma} adds additional adapter layers into frozen LLMs to augment them with visual capabilities. It also finetunes a visual encoder to transform images into visual prompts as prefixes.
    \item \textbf{BLIP-2}~\cite{Li:2023blip2} introduces a Querying Transformer to extract visual features from the frozen image encoder. It then utilizes these features as visual prompts to augment frozen LLMs with visual information.
\end{itemize}

We show the details of VaLMs evaluated on \method in Table~\ref{tab:valm-details}, including the extra model parameters (except the \textit{frozen} LLM backbone), the architecture of the visual encoder, and the number of pretraining images. Since the VaLMs vary in implementation details (e.g., the visual encoder), we cannot make a direct (head-to-head) comparison between the VaLMs and leave it for future investigation.

\section{Details of Main Experimental Results}
\label{sec:main-results-details}
We provide detailed experimental numbers of our main results with LLMs and VaLMs in Table~\ref{tab:scaling-results} and \ref{tab:valm-results}, respectively. For LLMs, we show the zero- and few-shot evaluation results of OPT, GPT, Pythia, Falcon, and LLaMA across various model scales. For VaLMs, we compare the performance of Z-LaVI, BLIP-2, and MAGMA with their \textit{frozen} LLM backbones.

\begin{table}[t]
\centering
\small
\begin{tabular}{l}
\toprule
\textit{\textbf{Prompt 1}} \\
\midrule
Answer List: {[}CANDIDATES{]}  \\
{[}QUESTION{]} Please select the most possible answer \\
from the above list. Please answer in one word. \\ \midrule
\textit{\textbf{Prompt 2}} \\ \midrule
Answer List: {[}CANDIDATES{]} \\
{[}QUESTION{]} Please only print the answer selected \\
in the above list. Please answer in one word.  \\ \midrule
\textit{\textbf{Prompt 3}} \\ \midrule
{[}QUESTION{]} Please select the most possible answer \\
from {[}CANDIDATES{]}. Please answer in one word. \\
\bottomrule
\end{tabular}
\caption{The prompts we utilize for ChatGPT evaluation on \method. ``{[}CANDIDATES{]}'' denotes the answer set of the evaluated subset in \method, as shown in Table~\ref{tab:answer-set}.}
\label{tab:chatgpt-prompts}
\end{table}

\section{Evaluation Details of Other Models}
\label{sec:other-eva-details}
This section provides evaluation details of VQA finetuned multimodal models and RLHF models.

\paragraph{Multimodal Models} For multimodal models such as BLIP, we adopt the evaluation settings used in VQAv2~\cite{Goyal:2017VQAv2}. Specifically, for each QA pair and its corresponding image, we evaluate the model using open-ended generation and obtain the output answer. Based on the experimental settings outlined in Section~\ref{sec:valm-eval}, we provide the top-10 ranked images for each QA pair and determine the final answer by majority prediction.

\paragraph{RLHF Models} We evaluate ChatGPT with constrained prompts and automatically compute the top-1 accuracy. The prompts utilized are presented in Table~\ref{tab:chatgpt-prompts}.

\end{document}